%% file: main.tex
\documentclass[runningheads]{llncs}

\usepackage{graphicx}
\usepackage{booktabs}
\usepackage{adjustbox}
\usepackage{multirow}
\usepackage{todonotes}
\usepackage[pagebackref]{hyperref}

\usepackage[accsupp]{axessibility}  

\usepackage{hyperref}
\usepackage{wrapfig}

\usepackage{orcidlink}

\begin{document}

\title{Backdoor Directions in Vision Transformers}

\author{Sengim Karayal\c cin\inst{1}\and
Marina Kr\v cek\inst{2}\and Pin-Yu Chen \inst{3}
Stjepan Picek\inst{4,2}}

\authorrunning{S. ~Karayal\c cin et al.}

\institute{Leiden University \and Radboud University \and IBM Research \and University of Zagreb Faculty of Electrical Engineering and Computing
\email{s.karayalcin@liacs.leidenuyniv.nl}\\
}

\maketitle

\begin{abstract}
This paper investigates how Backdoor Attacks are represented within Vision Transformers (ViTs). By assuming knowledge of the trigger, we identify a specific ``trigger direction'' in the model’s activations that corresponds to the internal representation of the trigger. We confirm the causal role of this linear direction by showing that interventions in both activation and parameter space consistently modulate the model's backdoor behavior across multiple datasets and attack types.

Using this direction as a diagnostic tool, we trace how backdoor features are processed across layers. Our analysis reveals distinct qualitative differences: static-patch triggers follow a different internal logic than stealthy, distributed triggers. We further examine the link between backdoors and adversarial attacks, specifically testing whether PGD-based perturbations (de-)activate the identified trigger mechanism. Finally, we propose a data-free, weight-based detection scheme for stealthy-trigger attacks. Our findings show that mechanistic interpretability offers a robust framework for diagnosing and addressing security vulnerabilities in computer vision.


\end{abstract}

\section{Introduction}
\label{sec:introduction}

Backdoor attacks are attacks against machine learning systems that aim to corrupt a small fraction of the training dataset with inputs containing a specific trigger pattern~\cite{DBLP:journals/access/GuLDG19}. The attacker's goal is then to make the model behave differently on inputs containing this trigger. Recent work has also shown that poisoning is a practical concern for web-scale image and text datasets~\cite{DBLP:conf/sp/CarliniJCPPATTT24,souly2025poisoning}.

Backdoor attacks have been very well studied in image classification models. The common setup is for the attacker to backdoor a fraction of images with a trigger, which then results in misclassifications to a specific target class~\cite{Trojannn,DBLP:journals/access/GuLDG19,li2022backdoor}. To defend against these attacks, a large number of defenses have also been explored~\cite{li2021invisible,bai2024backdoor,backdoorbench2025}.
However, while these defenses work well for convolutional models, their performance decreases significantly when deployed for Vision Transformer (ViT) models~\cite{backdoorbench2025}.\footnote{See BackdoorBench leaderboard at \url{https://backdoorbench.github.io/leaderboard-cifar100.html} for detailed defense results.}

As such, it is necessary to design ViT-specific defenses. However, initial attempts show relatively poor performance across unknown attack scenarios. While specific defenses have been proposed~\cite{doan2023defending,subramanya2024closer}, they focus on detecting anomalous attention patterns and do not perform effectively against more distributed triggers. This fragility of ViT-specific defenses to standard backdoor attacks highlights our limited understanding of how these networks represent and propagate backdoor features internally.\\

\begin{wrapfigure}{r}{0.6\linewidth}
    \includegraphics[width=0.9\linewidth]{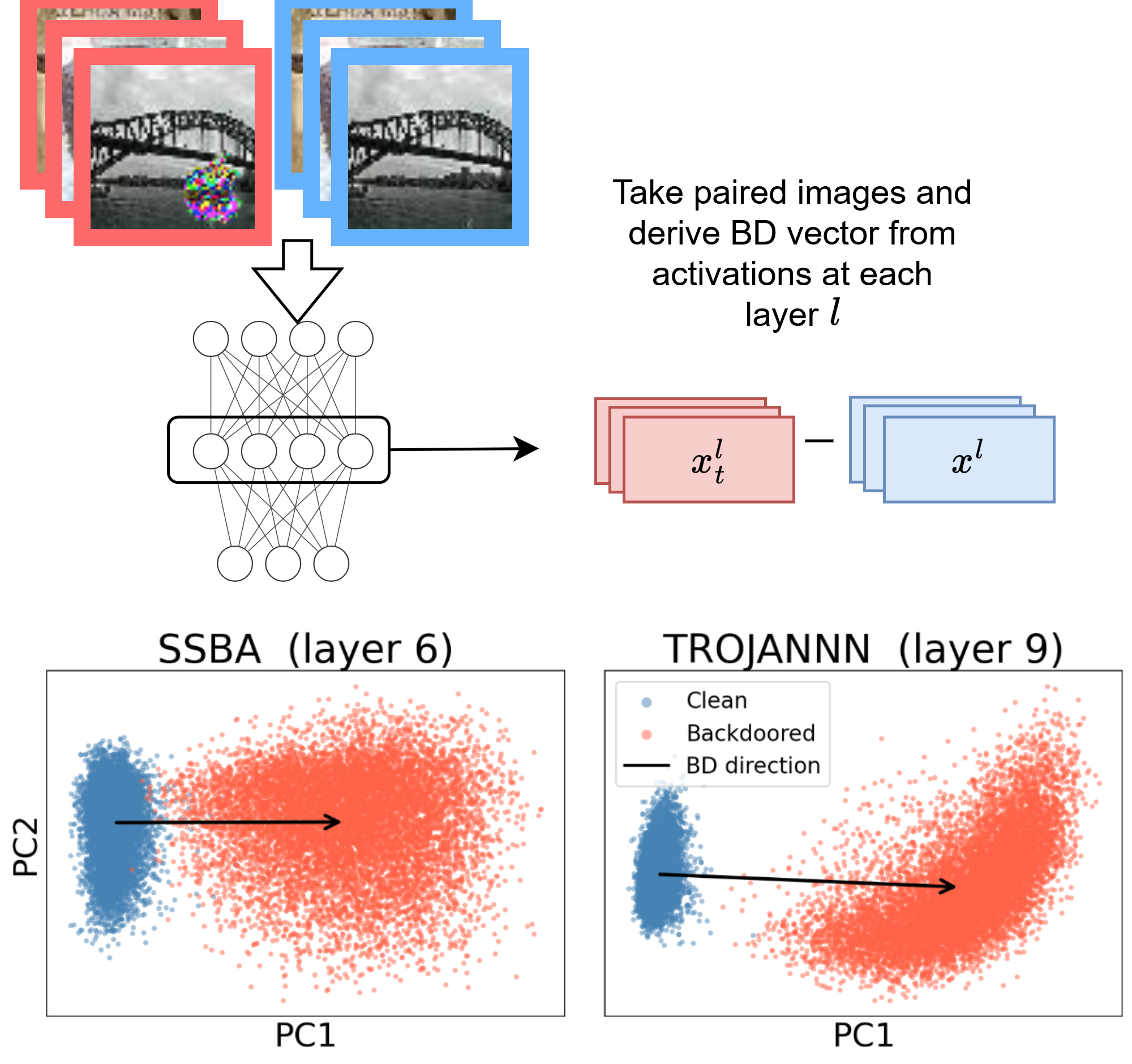}
    \caption{Derivation of the backdoor (BD) direction at layer $l$ (top). PCA projections of internal activations for clean and backdoored images at an intermediate layer alongside the resulting BD direction (bottom).}
    \label{fig:abstr}
\end{wrapfigure} 

Motivated by recent advances in understanding how transformer-based language models represent concepts using linear directions~\cite{park2023lrh,rimsky-etal-2024-steering}, we aim to analyze the underlying representation of backdoor attacks in ViTs by utilizing some of those advanced techniques.
The main goal is to understand better where in the model the trigger is detected and how different the mechanisms are behind various attacks. In our view, a better understanding of how models perform these computations will help design more robust defenses. 

Concretely, we assume full knowledge of the backdoor trigger and the backdoored training data, as this enables more detailed analyses across multiple datasets and attacks. For the initial analyses, we follow the setup in~\cite{refusal}, where the authors show the refusal mechanism in language models is mediated by a single direction in the activation space (see Figure~\ref{fig:abstr}). Notably, we show that similar analyses allow us to define a backdoor direction for each model. We then verify that this direction causally influences model outputs by (de-)activating the backdoor using activation steering. Additionally, we show that orthogonalizing this direction out of weight matrices results in the removal of the backdoor from the model.

After validating that a single direction can modulate backdoor behavior across attacks and datasets, we leverage knowledge of this direction to analyze layer-wise propagation of trigger information to understand differences between trigger types. Subsequently, we use this direction to provide insights into how adversarial examples interact with backdoored models internally. This is motivated by several recent works that have shown that PGD-based attacks behave differently on clean vs. backdoored models~\cite{DBLP:journals/pami/NiuSMJH24,DBLP:conf/aaai/YinWLLL25}. Finally, we propose a scheme to detect backdoored models purely based on weights. This is similar to previous methods for weight-based backdoor detection based on the Lipschitzness of the channels in convolutional layers~\cite{zheng2022data,lipshits2025}. However, these schemes are tailored to convolutional models and do not directly translate to ViTs.

Our main contributions can be summarized as follows:
\begin{enumerate}
    \item Using knowledge of the trigger, we obtain the backdoor direction in backdoored models and show that removing this direction from weights mitigates the backdoor, confirming its causal role in the model's behavior.
    \item We show that the layers in which models learn the same trigger pattern often behave similarly across poisoning rates and datasets.
    \item We use the derived direction to provide additional insights into the internal mechanisms underlying the connection between backdoored models and adversarial examples as found in~\cite{DBLP:journals/pami/NiuSMJH24,DBLP:conf/aaai/YinWLLL25}.
    \item We show that ViT's architecture enables relatively simple weight-based backdoor detection for stealthy backdoor attacks such as WaNet and BPP.
\end{enumerate}


\section{Preliminaries}
\label{sec:preliminaries}

\subsection{Backdoor Attacks}
Data backdoor attacks, where an adversary implants a trigger into training data so that a model learns to respond to that trigger at test time, are a well-studied threat to modern computer-vision systems. We can divide common trigger patterns into several categories: patch-based triggers, where a localized patch is introduced in images~\cite{DBLP:journals/access/GuLDG19,Trojannn}; blended triggers, where a fixed pattern is mixed into the image~\cite{blended}; or dynamic triggers, where the trigger pattern is included in images in an input-dependent way~\cite{nguyen2021wanet,li2021invisible,wang2022bppattack}. 

We can also divide attacks based on the attacker's goals and capabilities. In targeted attacks, the goal of the attacker is to make the model predict a specific target class when the trigger is present~\cite{DBLP:journals/access/GuLDG19}. Conversely, untargeted attacks aim to cause incorrect predictions in general~\cite{luo2023untargeted}.
The labeling mechanism can be divided into \textbf{dirty-label} attacks, where the attacker can both introduce triggers to images and change their labels~\cite{DBLP:journals/access/GuLDG19}, and \textbf{clean-label} attacks, where the attacker can only introduce triggers while keeping the original labels unchanged~\cite{turner2019label}. In this work, we only consider targeted, dirty-label attacks.

\subsection{Transformer Interpretability}
A Transformer~\cite{vaswani2017attention} consists of an embedding layer, stacked Transformer blocks, and an output head. Inputs are mapped to embeddings and processed by transformer blocks combining an attention and MLP layer, and a residual shortcut connection. We will refer to this as the residual stream.
In Vision Transformers (ViTs)~\cite{vit2021reference}, images are split into fixed-size patches that serve as tokens. Additionally, a special \texttt{[CLS]} token is added to collect global information, and its final representation is used for prediction.

Following the mechanistic interpretability view~\cite{elhage2021mathematical}, attention moves information between tokens, while MLPs act on individual tokens. The residual stream is the central state to which all components apply linear read–write operations. We view it as a $d$-dimensional space where the representation is a linear combination of many feature directions, i.e., directions representing features relevant for the learned tasks~\cite{park2023lrh}. Note that the number of feature-directions learned by large models can be larger than $d$~\cite{elhage2022superposition}, so features do not need to be linearly independent.

Directions in the residual stream, corresponding to human-interpretable features, can be derived using contrastive sets/pairs of examples (positive vs. negative for a feature)~\cite{rimsky-etal-2024-steering} or unsupervised methods such as sparse autoencoders~\cite{bricken2023monosemanticity}. Identified directions enable activation steering, where they are added to the residual stream during a forward pass to control model behavior~\cite{rimsky-etal-2024-steering}, and can also guide weight updates to suppress specific features~\cite{refusal}.

\section{Related Work}
\label{sec:related}

\paragraph{Backdoors on ViTs.}
Since the introduction of ViTs~\cite{vit2021reference}, there have been several explorations into ViT-specific attacks. BadViT aims to ensure higher attention scores for the patch containing the trigger~\cite{yuan2023badvit}. TrojViT aims to insert backdoors that can be activated by flipping a limited number of bits in the weights~\cite{zheng2023trojvit}. In~\cite{doan2023defending,subramanya2024closer}, two attention-based defenses are proposed that aim to detect outliers in the attention maps to detect backdoored examples, although these only work well for patch-based triggers. Next, in~\cite{wang2025attention,gong2025megatron}, adaptive attacks are proposed that avoid such detection-based defenses. Finally, in~\cite{li2024using}, an unlearning-based defense is proposed for ViTs. 

\paragraph{Mechanistic Interpretability in Vision Models.}
While recent work in the field of mechanistic interpretability has focused on LLMs, some of the foundational works on mechanistic interpretability focused on finding interpretable neurons in convolutional vision models~\cite{olah2017feature,olah2018building,olah2020zoom}. More recently, some works applied sparse autoencoders to InceptionV1 models to find additional curve-detectors~\cite{gorton2024the}, and to evaluate the relationship between superposition and adversarial examples~\cite{gorton2025adversarial}. Furthermore, there are many works on interpreting multi-modal foundation models~\cite{lin2025survey}. Finally, in~\cite{gupta2025no}, activation steering is used to correct for biases in ViT-based image classifiers without retraining.

In the context of backdoors, we can view feature-space trigger inversion approaches~\cite{wang2023unicorn,xu2024towards,xu2024ban} as interpretability methods that aim to find a backdoor feature in the network. However, these techniques generally focus on finding a set of backdoor neurons, which is less principled in transformer models due to features not necessarily being aligned with the standard basis in the transformer residual stream~\cite{elhage2021mathematical,elhage2022superposition}. Additionally, these techniques focus on the extracted features, which ignore intermediate layers.

\paragraph{Adversarial Examples and Backdoors.}
The connection between adversarial examples and backdoor attacks has been drawn repeatedly. From the perspective that adversarial examples result from non-robust features, i.e., superficial correlations in the training data~\cite{ilyas2019features}, we can view backdoor triggers as a special case of such non-robust features. Indeed, several works have explored how adversarial examples behave against backdoored models. First, in~\cite{weng2020tradeoffs}, the authors showed a tradeoff between adversarial robustness and vulnerability to backdoor attacks. Furthermore, in~\cite{wei2023shared}, it is found that unlearning shared adversarial examples can mitigate backdoors. In~\cite{DBLP:journals/pami/NiuSMJH24}, the authors showed that adversarial examples against backdoored models tend to result in misclassifications to the target class. Similarly, in~\cite{DBLP:conf/aaai/YinWLLL25}, the authors demonstrated that adversarial attacks starting from backdoored examples tend to result in `misclassifications' to the original clean class. Finally, in~\cite{min2024uncovering}, it is shown that adversarial examples against purified models can reactivate the backdoor.

Overall, the theme of these works is that generating adversarial examples against backdoored models can serve as a defense against them. However, these works are limited to convolutional models. Only in~\cite{DBLP:journals/pami/NiuSMJH24} authors assess the similarity between generated adversarial examples and backdoored images in the extracted features, but this again ignores the similarity throughout the rest of the network. 

\section{Tested Models}
In our experiments, we use the CIFAR10, CIFAR100, and TinyImageNet datasets because they offer a comprehensive range of complexity, from standard benchmarks (like CIFAR10) to more challenging datasets that are essential for evaluating the scalability of our methods on higher-resolution images and a larger number of classes.
Regarding the investigated models, we use pretrained models from the BackdoorBench~\cite{backdoorbench,backdoorbench2025} list to reduce the computational cost of training new models, since our method is a post-training analysis. Moreover, using these public, established models eliminates the training bias introduced by (our) specific training choices and emphasizes the method's applicability. The models we consider are standard ViT-B-16~\cite{vit2021reference} models with a patch size of 16 and 12 encoder blocks. For each dataset and attack, we use models trained with three poisoning rates $\{0.01, 0.05, 0.1\}$, as lower poisoning rates generally result in unsuccessful attacks. Note that at a poisoning rate $0.01$, some of the tested models only reach around 40-70\% attack success rate (ASR). 
We use WaNet~\cite{nguyen2021wanet}, SSBA~\cite{li2021invisible}, and TrojanNN~\cite{Trojannn} for all datasets, with the addition of BadNet~\cite{DBLP:journals/access/GuLDG19}, Blended~\cite{blended}, and BPP~\cite{wang2022bppattack} for CIFAR10. Finally, we add InputAware~\cite{nguyen2020input} for CIFAR100. These selections are based on the availability of models for download per dataset.\footnote{Note that we exclude low-frequency (LF) as the ASRs in available ViTs are always under 5\%.}

\section{Validating Linear Backdoor Direction}
\label{sec:validate}
We first focus on validating that there is a linear direction that describes the backdoor. We aim to identify a direction by taking contrastive pairs of clean and backdoored images. Then, we can use this direction to directly intervene in the activation space during the forward pass, and in weights directly updating models to remove the backdoor. 

We note that defining the direction by taking contrastive pairs requires full knowledge of the backdoor trigger. As such, while we show performance tables with reductions in ASRs, this methodology is not intended as a defense, as a priori knowledge of the trigger is an unrealistic assumption. 

\subsection{Defining the Direction}
Let $X$ represent the original clean training dataset. We select a subset of $X$, denoted $X_{s} \subseteq X$, to apply backdoor triggers. We define the backdoored sample corresponding to $x \in X_s$ as $x_t$.
Then, we have a pair of clean and backdoored images  $(x, x_t)$, where $x_t$ is the backdoored version of $x$, and a set: $X_{\text{pair}} = \{(x, x_t) \mid x \in X_s\}$ (see Figure~\ref{fig:abstr}).
For a given layer $l$, define the activations of an input $x$ as $x^l \in \mathbb{R}^d$, where $d$ is the dimension of the flattened intermediate representation.
To define a direction $r^l$, we take the averaged difference vector between backdoored and clean images to find the intermediate representation of the trigger:
\[
    r^l = \frac{1}{|X_{\text{pair}}|}\sum_{(x, x_t) \in X_{\text{pair}}} (x^l_t - x^l).
\]
\\
We define two types of activation vectors for analysis:
\begin{itemize}
    \item The vector corresponding to only the \texttt{[CLS]} token.
    \item The vector obtained by concatenating all tokens, including the image and \texttt{[CLS]} token.
\end{itemize}

\subsection{Experiments}
We validate the identified backdoor direction through activation steering and weight orthogonalization. By observing how these interventions affect the success (or removal) of the backdoor behavior, we can confirm the direction's causal significance.
\paragraph{Activation Steering.}
We use additive activation steering as an intervention~\cite{rimsky-etal-2024-steering}, where we add the vector (found backdoor direction) to clean images and subtract it from backdoored images. We measure the efficacy of these interventions by taking ASR and recovered accuracy (RA) for positive and negative steering.

To assess the effects of (de-)activating the backdoor feature in specific layers, we steer using the found directions during the forward pass. Steering in layer $l$ with vector $r^l$ is then realized by adding $r^l$ to activations $x^l$ during the forward pass. To assess how well the directions describe backdoor behavior, we do two main interventions:
\begin{itemize}
    \item \textbf{Positive Steering:} For a clean test set $X_c$, we add the corresponding $r^l$ at each layer $l$. Then, we define the attack success rate for this vector, $ASR_{+r^l}$, as the fraction of images that are flipped to the target class.
    \item \textbf{Negative Steering:} For a test set containing only backdoored images $X_t$, we subtract the $r^l$. Then, we define the robust accuracy of the vector $r^l$ as $RA_{-r^l}$, the fraction of images that are classified to their original label.
\end{itemize}

We then define a most representative layer $\hat{r}$ (Eq.~\eqref{eq:backdoor_direction}) for the backdoor direction to have a single direction and layer for steering results in this section. Results analyzing the steering results across layers are in Section~\ref{sec:steering}.
\begin{equation}
\label{eq:backdoor_direction}
    \hat{r} = \text{arg max}_{r^l}\{(ASR_{+r^l} + RA_{-r^l}) - (ASR_{+r^{l-1}} + RA_{-r^{l-1}})\}.
\end{equation}

\paragraph{Weight Orthogonalization.}
\begin{table}[tb]
\centering
\caption{Average Steering and Orthogonalization for $\hat{r}$ results by dataset (excluding models with baseline ASR $< 0.9$).}
\label{tab:steer_by_dataset}
\begin{tabular}{lc|ccc|cc|cc|ccc}
\toprule
 & $N$ & \multicolumn{3}{c|}{Baseline} & \multicolumn{2}{c|}{Steering CLS} & \multicolumn{2}{c|}{Steering All} & \multicolumn{3}{c}{Orthog.} \\
 &  & ASR & RA & CA & $ASR_{+\hat{r}}$ & $RA_{-\hat{r}}$ & $ASR_{+\hat{r}}$ & $RA_{-\hat{r}}$ & ASR & RA & CA \\
\midrule
CIFAR10 & 12 & 97.6 & 2.3 & 95.4 & 59.3 & 22.8 & 96.5 & 70.6 & 0.3 & 84.6 & 93.9 \\
CIFAR100 & 13 & 96.9 & 2.6 & 78.6 & 62.1 & 19.7 & 87.5 & 39.5 & 15.9 & 51.4 & 78.0 \\
TINY & 8 & 99.0 & 0.8 & 70.7 & 89.0 & 21.6 & 99.5 & 51.9 & 1.3 & 56.4 & 70.4 \\
\midrule
\textbf{Overall} & 33 & 97.7 & 2.1 & 82.8 & 67.6 & 21.3 & 93.7 & 53.8 & 6.7 & 64.7 & 82.0 \\
\bottomrule
\end{tabular}
\end{table}
Following~\cite[Section 4.1]{refusal}, a behavior can be removed by orthogonalizing the vector associated with that behavior at each layer that contributes to the residual stream. This is then a useful way of testing whether a single direction is responsible for the backdoor in ViTs. Consequently, we update the initial embedding layer and all attention and MLP output projection matrices. The new weights are then defined as: 
\[
    \mathbf{W_{new}} = \mathbf{W} - \hat{r}\hat{r}^T\mathbf{W}.
\]

\paragraph{Results.}
In Table~\ref{tab:steer_by_dataset}, we can see summarized results for both steering and orthogonalization. Overall, for steering, both CLS and all-token steering with the found direction can both amplify and mitigate the backdoor to some degree. However, especially the CLS-only steering does not always fully increase ASR or increase RA to better levels. While these results suggest that steering is not particularly effective, the effects of steering with a specific direction are often highly sensitive to specific parameters~\cite{rimsky-etal-2024-steering,refusal}, and we refrain from tuning these parameters or steering across multiple layers as our goal is not to optimize steering performance. A more detailed analysis of layer-wise behavior can be found in Section~\ref{sec:steering}.

When we examine the overall orthogonalization results, we see that removing the backdoor direction from weights almost always eliminates the backdoor. The only attack where it does not is the blended attack on CIFAR100, whereas in every other attack, ASR is reduced to below 5 while only minimally affecting clean accuracy. While the recovered accuracy is not the same as the clean accuracy, this difference can be (at least partially) attributed to image distortion caused by the trigger. Overall, both the steering and orthogonalization results indicate that for each model, a single linear direction in the models' residual stream modulates backdoor behavior. 

\section{Layer-wise Steering}
\label{sec:steering}

In this section, we will use the derived directions to analyze layer-wise propagation of the backdoor trigger. We show a subset of models due to space constraints. Results for all models are available in the supplementary material. 


\begin{figure}[tb]
    \centering
    \includegraphics[width=0.9\linewidth]{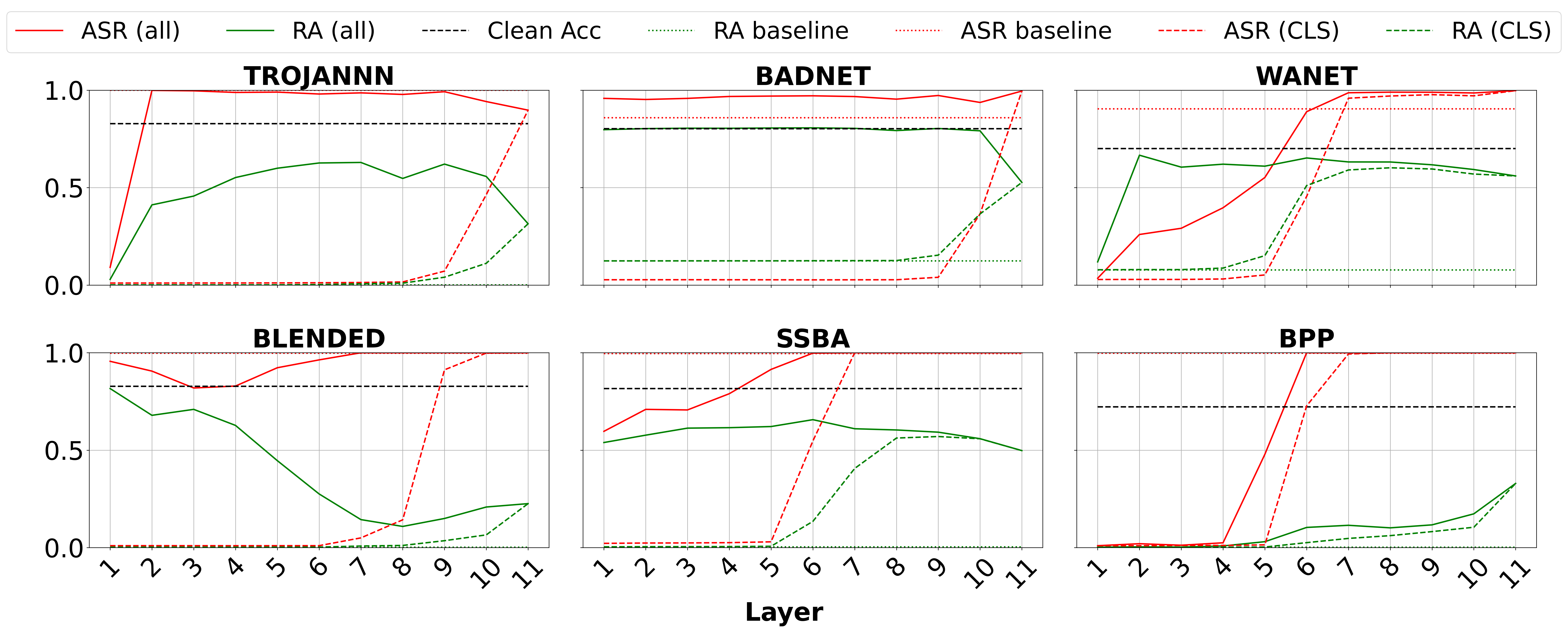}
    \caption{Steering results on CIFAR100 with 0.1 poisoning rate.}
    \label{fig:steering_pr_01_cifar100}
\end{figure}

\subsection{Varying Attacks}
In Figure~\ref{fig:steering_pr_01_cifar100}, for attacks with static triggers, the vector defined over all tokens works best in the earliest layers. Steering positively results in high ASRs for early layers. This indicates the earliest layers have not yet moved trigger-related information across patches, and that the representation of different parts of the trigger is not unified.
To illustrate, detecting the \texttt{hello-kitty} trigger for the Blended attacks is different in different token positions. The trigger is not represented with the same feature across different parts of the image, and the trigger is a set pattern that covers the entire image. Accordingly, detecting the trigger requires the model to find different features in different patches. This also applies to attacks like BadNet and TrojanNN that have static triggers covering a specific part of the image. 
For more dynamic triggers like InputAware, WaNet, and BPP, this is less clear. As these triggers vary for each input, steering all tokens with a static vector does not reach baseline ASR in earlier layers.

When we consider only the \texttt{[CLS]} token, we see that in early layers, it generally does not move ASR or RA. Generally, there is a sharp increase in ASR between 2-3 layers, but which layers those are varies per attack. 
This indicates that from these layers on, the $r^l$ is actually representing the trigger in the \texttt{[CLS]} token.
We still often see that RA can increase after these layers. While this may indicate that some backdoor activity remains in other tokens, another option is that in later layers, subtracting $r^l$ might result in the model never predicting the target class.

\begin{figure}[t]
    \centering
    \includegraphics[width=0.9\linewidth]{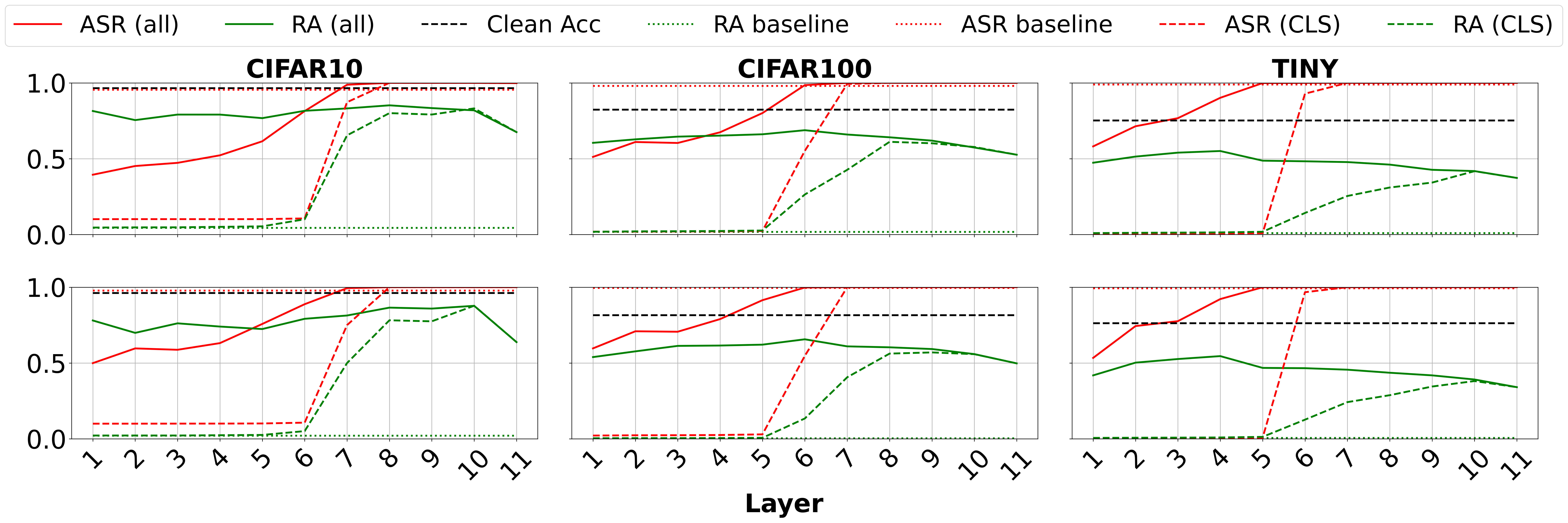}
    \caption{Steering SSBA across datasets and poisoning rates (0.05 top, 0.1 bottom).}
    \label{fig:ssba_steering_pr}
\end{figure}

\subsection{Varying Dataset and Poisoning rates}
In Figure~\ref{fig:ssba_steering_pr}, we showcase results for SSBA across datasets and poisoning rates. As we can see, the shapes of the curves are similar for all models and datasets. The exact layer for the initial ASR increase does not fully match (layer 6 for CIFAR10, layer 5 for others), but we still see the RA starts increasing in layer 5 for CIFAR10.
Note that we exclude plots for a poisoning rate of 0.01 due to ASRs being lower. These graphs look marginally different, though we still see ASR and RA in the \texttt{[CLS]} token start increasing from layer 5-6. We attribute the difference primarily to the attacks not being entirely successful against these models ($ASR < 0.95$). Importantly, this consistent layer-wise behavior across datasets we see for SSBA also holds for the other attacks we tested (see supplementary material).

\subsection{Generality for other ViT Variants}
\begin{wrapfigure}{r}{0.5\linewidth}
    \centering
    \includegraphics[width=0.95\linewidth]{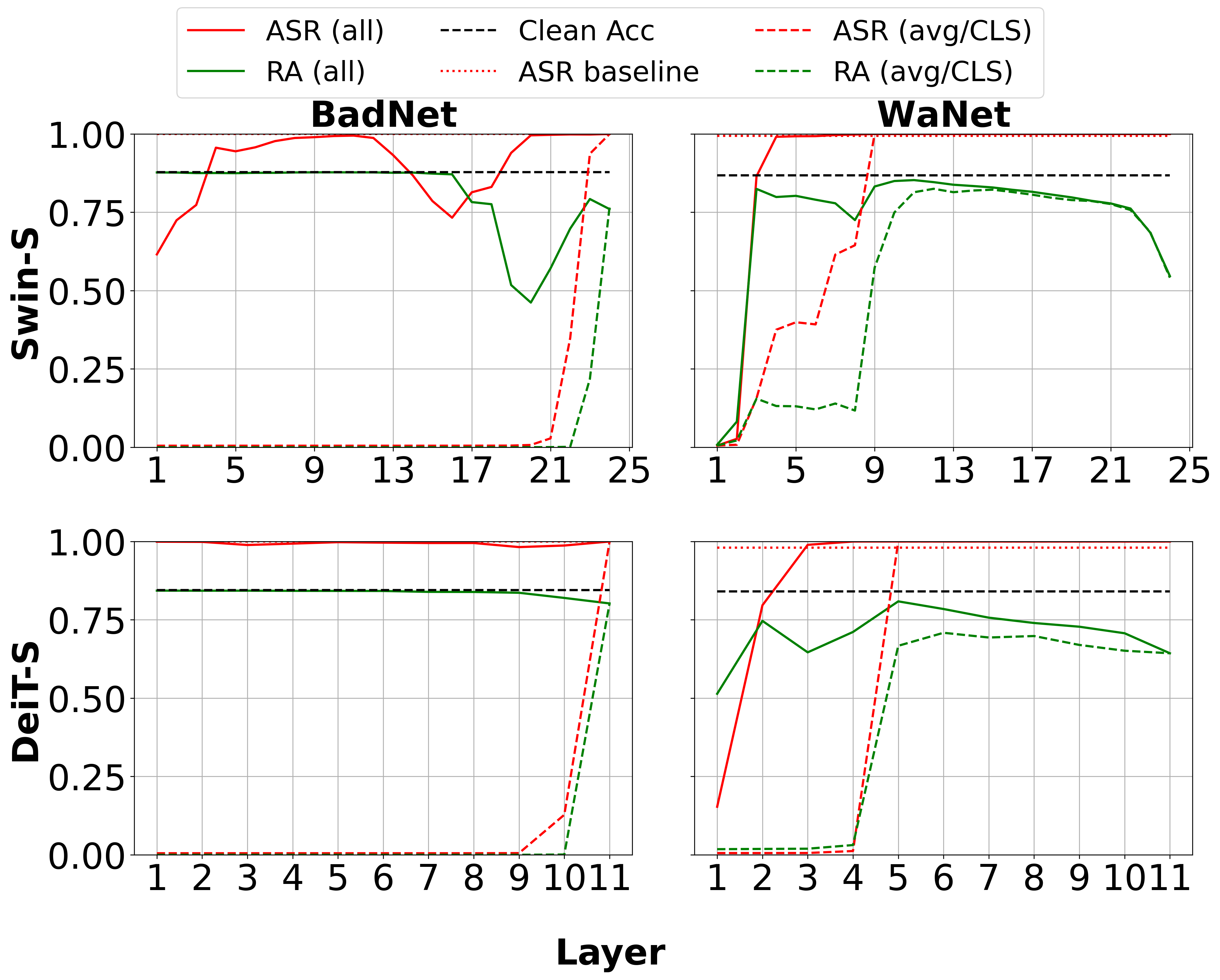}
    \caption{Backdoor steering for Tiny-ImageNet for DeiT-S/Swin-S}
    \label{fig:rebuttal}
\end{wrapfigure}
To assess the generality of these takeaways, we also consider some different ViT variants. We finetune DeiT-S and Swin-S models on Tiny-Imagenet for BadNet and WaNet. We use a poisoning ratio of 0.1, and finetune pre-trained weights for 5 and 20 epochs for BadNet and WaNet, respectively, to obtain sufficiently high ASRs. Note that as Swin does not have a \texttt{[CLS]} token, we consider the average vector across tokens.

We see similar conclusions as for earlier models. For BadNet, the all-token steering works from the earliest layers, while \texttt{[CLS]} steering only works in the final few layers. Conversely, for WaNet, we see that all-token steering only works from a few layers in, while \texttt{[CLS]} steering starts working from relatively middle layers.

\section{Adversarial Examples and Backdoors}
\label{sec:adversarial}
\begin{table}[b]
\centering
\caption{Percentage of AE Predictions for Target/Original Class - CIFAR100}
\label{tab:percentage_bd}
\begin{tabular}{l| ccc|ccc}
\toprule
Attack & \multicolumn{3}{c}{Clean}  & \multicolumn{3}{c}{Backdoor} \\
 & 0.01 & 0.05 & 0.1 & 0.01 & 0.05 & 0.1 \\
\midrule
BADNET & 0.6\% & 5.6\% & 7.7\% & 33.2\% & 32.9\% & 33.0\% \\
BLENDED & 0.2\% & 0.2\% & 0.2\% & 25.0\% & 22.2\% & 22.3\% \\
WANET & 1.9\% & 1.9\% & 4.1\% & 28.9\% & 25.1\% & 26.0\% \\
SSBA & 0.2\% & 0.0\% & 0.1\% & 26.2\% & 25.1\% & 25.9\% \\
BPP & 29.2\% & 31.4\% & 41.5\% & 18.7\% & 21.7\% & 21.9\% \\
TROJANNN & 0.6\% & 0.8\% & 0.7\% & 35.2\% & 45.9\% & 47.6\% \\
\bottomrule
\end{tabular}
\end{table}
In this section, we investigate how adversarial examples (AEs) interact with backdoored models. In~\cite{DBLP:journals/pami/NiuSMJH24}, we see that for backdoored models, adversarial perturbations on clean images disproportionately result in misclassifications to the target class. Similarly, in~\cite{DBLP:conf/aaai/YinWLLL25}, we see that when starting from backdoored images, generating adversarial examples often results in the original clean class. Additionally, the number of PGD steps required to flip the label of a backdoored image is substantially larger than for clean images~\cite{DBLP:conf/aaai/YinWLLL25}. In both works, this insight is leveraged to create backdoor defenses.

The results in these papers indicate that the distribution of output classes can indicate which class could be backdoored~\cite{DBLP:journals/pami/NiuSMJH24}. Moreover, the additional steps required to generate adversarial perturbations for backdoored images can help identify backdoored examples as in~\cite{DBLP:conf/aaai/YinWLLL25}. 


\paragraph{Experimental Setup.} For both subsections, we use PGD with $l_{\infty}$-norm and $\epsilon = 8/255$. We start from either clean or backdoored test images and run 5 or 15 steps for clean and backdoored examples, respectively.

Then, we obtain pairs of $(x_i, a_i)$, where $a_i$ is the perturbed image, and the class $y_{a_i}$ that $a_i$ is now predicted as. We can calculate the differences in activations for each of these at every layer by computing $v^l_i = a^l_i - x_i^l$. Here, we focus only on the \texttt{[CLS]} token, as analyzing the all-token vector is more difficult due to its high dimensionality.

\subsection{Do AEs Use Backdoor Features?}
\label{sec:start_clean_ae}

When we first examine the percentage of perturbed images that get classified as the target class in Table~\ref{tab:percentage_bd}, we clearly see that BPP results in the most misclassifications to the target class. At higher poisoning rates, BadNet and WaNet also reach around 5\% of images. However, for these models, there are other classes that get `targeted' by PGD with similar percentages. The results for Tiny-ImageNet show similar trends, while results for CIFAR10 are more aligned with~\cite{DBLP:journals/pami/NiuSMJH24}. These results are in the supplementary material.

\begin{figure}[!t]
    \centering
    \includegraphics[width=0.8\linewidth]{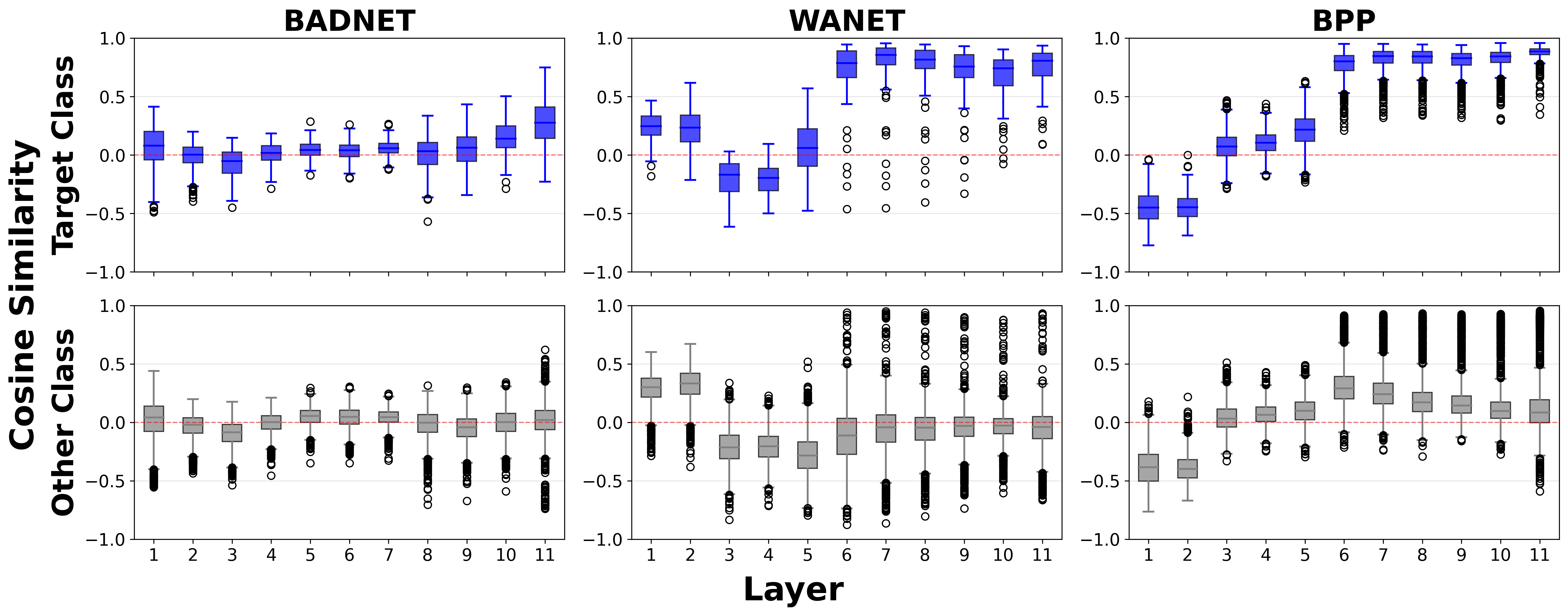}
    \caption{Distribution of cosine similarities across layers for CIFAR100 with poisoning rate 0.05, starting from clean images}
    \label{fig:cosine_sims}
\end{figure}

Next, we investigate whether the adversarial examples actually use the backdoor direction. To do this, in each layer $l$, we take the differences $v^l_i$ and compute the cosine similarity with the backdoor vector $r^l$.
We can compare the distributions of cosine similarities for images that are flipped to the target class and those that are not. In Figure~\ref{fig:cosine_sims}, we showcase examples of these distributions for attacks where a larger percentage of examples go to the target class. For WaNet and BPP, there is a clear trend towards higher cosine similarities in middle layers. For BadNet, this is not the case. Only in the final layers is the similarity higher, but as it is still substantially lower than for the other attacks, we mostly attribute this to these examples being predicted as the target class. This matches the intuition that for stealthier attacks, where the triggers are not visible, they can be exploited by PGD, which is restricted by the $\epsilon$ parameter to generate perturbations within an imperceptible range.


For a more comprehensive view, we present the cosine similarity distributions across all layers and attacks in tables in the supplementary material. Generally, the adversarial example vectors follow similar distributions to BadNet in Figure~\ref{fig:cosine_sims} (except for BPP and WaNet). These trends tend to hold overall for all attacks and datasets. Note that for Tiny-ImageNet, we also see that SSBA belongs to the attacks that get exploited by PGD. 

\subsection{Do AEs Remove Backdoor Features?}
\label{sec:start_bd_ae}

When we start from backdoored images, Table~\ref{tab:percentage_bd} shows that the new prediction for 20-50\% of backdoored images is their original class. This trend also holds for other poisoning rates and datasets.
\begin{figure}[!b]
    \centering
    \includegraphics[width=0.8\linewidth]{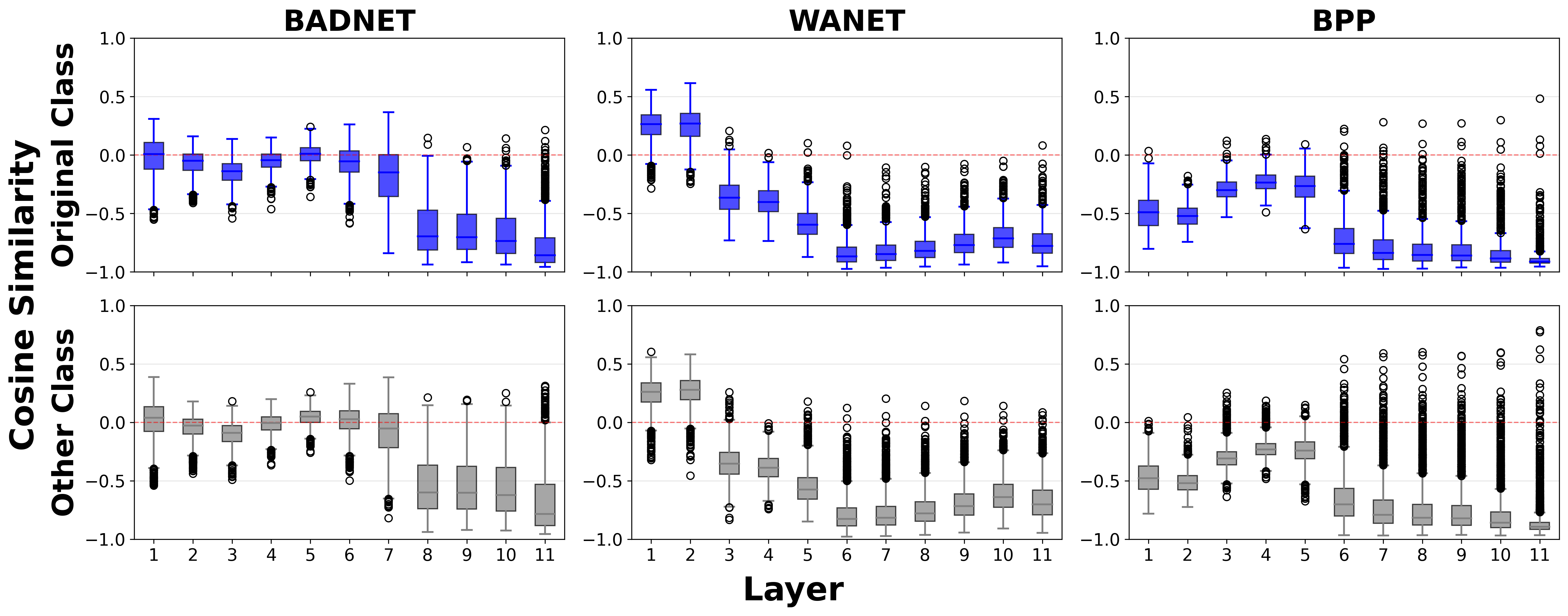}
    \caption{Distribution of cosine similarities across layers for CIFAR100 with poisoning rate 0.05, starting from backdoored images.}
    \label{fig:cosine_revs}
\end{figure}

In Figure~\ref{fig:cosine_revs}, the (negative) cosine similarity is consistently high in the later layers for all attacks. Furthermore, even for images where the predicted class does not become the original label, the cosine similarity is often $< -0.5$. Overall, there does not seem to be a significant difference between images that are or are not classified to the original class. These trends also hold for the other attacks, datasets, and poisoning rates. Complete results are in the supplementary material.

\section{Detecting Backdoors from Weights}
\label{sec:detecting}

In this section, we investigate whether we can leverage intuitions about transformer computation to design a weight-based detection defense. This approach is based on the idea that shortcut connections induced by the backdoor might result in signatures in early output projection weights (i.e., the weights from which we remove the backdoor direction using the orthogonalization). 

As such, we take the classifier head matrix \( O \in \mathbb{R}^{n_{class} \times d} \), where each row \( c_i \) represents the readout direction for class \( i \).
To identify potential backdoors, we measure the degree of alignment between each \( c_i \) and the layer weights \( W \) in the first $n$ layers.  
For each class \( i \), we compute a detection score $s_i = \sum_l \mathbb{I}\!\left[\text{abs}(c_i^\top W)_l > t\right]$ where \( t \) is a threshold and \( \mathbb{I}[\cdot] \) denotes the indicator function, i.e., we add any value in $\text{abs}(c_i^\top W)$ that is larger than $t$ to the score.

To detect whether the top class is an outlier, we also evaluate whether its score is an outlier. To do this, we define a $Z$-score:
\[Z = \frac{s_{top} - s_{second}}{\text{max}(\text{std}(S \setminus \{s_{top}\}),t)},\]
where $s_{top}$ and $s_{second}$ are the top two scores and $S = \{s_0,\cdots s_{n_{class}}\}$. We take the maximum of this standard deviation and the threshold for cases where the standard deviation is 0, which occurs when the top class is the only one to ever exceed $t$. We then define an outlier as $Z > 3$. 

\begin{figure}[!t]
    \centering
    \includegraphics[width=0.95\linewidth]{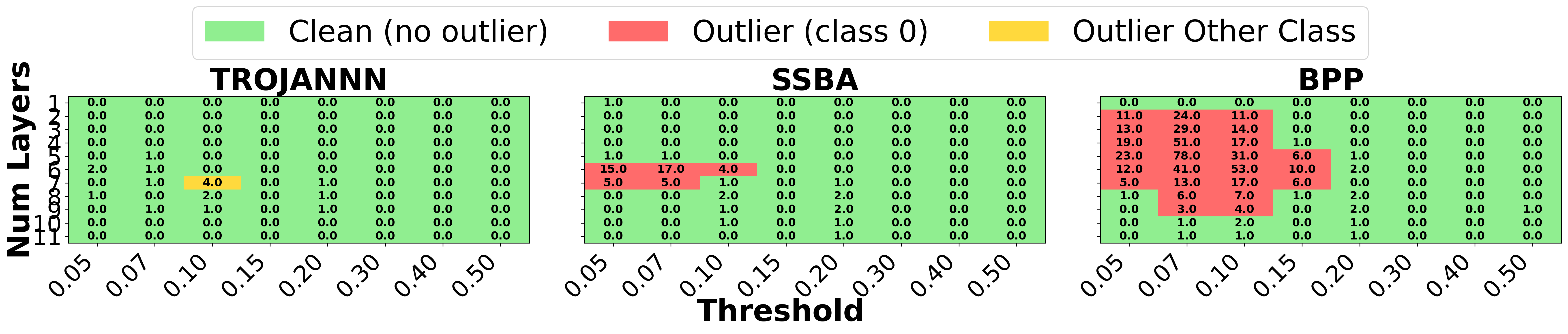}
    \caption{$Z$-scores for CIFAR100 models at poisoning rate 0.05.}
    \label{fig:grids_detection}
\end{figure}
In Figure~\ref{fig:grids_detection}, we show examples of grid searches for varying numbers of layers and thresholds. WaNet achieves a high score for the correct class across many tested combinations, whereas for SSBA, this fraction is smaller. For TrojanNN, we see no clear indication that the model is backdoored. 
The three different patterns are roughly consistent across all datasets and poisoning rates, which can be found in the supplementary material. Overall, WaNet and BPP are easy to detect, SSBA is often an edge case, while for attacks with patch triggers like TrojanNN, it does not work at all.

\section{Discussion}
\label{sec:discussion}

Our results clearly show that deriving a direction corresponding to the backdoor feature in the model is straightforward with the assumed knowledge of the trigger. This is not a surprising finding, given previous work showing that transformers represent features linearly~\cite{park2023lrh} and that we can derive a direction using contrastive pairs/sets~\cite{rimsky-etal-2024-steering}. Linear-directions corresponding to a large variety of concepts have been found in (vision) language models~\cite{bricken2023monosemanticity}. As such, confirming that this also holds for backdoor attacks in ViTs is a natural extension. It illustrates that backdoor models provide a convenient setting for studying and benchmarking mechanistic interpretability, since the relevant feature and its intended effect are clearly defined.

The results in Section~\ref{sec:steering} show some broader trends in how different attacks behave. First, for stealthier attacks (WaNet, SSBA, and BPP), the backdoor direction is moved to the \texttt{[CLS]} token in earlier layers than for those with static triggers.
The reason is that, in these attacks, triggers introduce minor perturbations that can be detected in each token separately. Conversely, other triggers will require more position-dependent detection mechanisms. 

Another observation is that the same attack generally results in fairly similar layer-wise behavior. This indicates that it might be feasible to design defenses that explicitly target specific types of trigger mechanisms. Indeed, our detection results in Section~\ref{sec:detecting} show that leveraging the early layer patterns for stealthier attacks allows lightweight detection of the backdoor model and target class.

\paragraph{On Adversarial Example Experiments.}
We observe that, when starting from clean images, the backdoor target class is not necessarily the most common target for adversarial examples, indicating that these trends are largely confined to datasets with fewer classes~\cite{DBLP:journals/pami/NiuSMJH24}. In particular, for attacks with human-perceptible triggers, this trend largely does not hold. For more stealthy trigger patterns, however, the difference vector for adversarial examples misclassified to the target class tends to be relatively similar to the backdoor direction in relatively early layers. Note that frequent misclassifications into a specific other class would complicate this analysis.\footnote{Which happens frequently, see supplementary materials for class-distributions.} However, finding common $v^l$ directions towards other classes could provide insights about finding non-robust features~\cite{ilyas2019features}.

When starting from backdoored images, we find that class trends are somewhat weaker than those reported in~\cite{DBLP:conf/aaai/YinWLLL25}, although the plurality of images still revert to their original class. While~\cite{DBLP:conf/aaai/YinWLLL25} already shows that more PGD steps are required to find adversarial examples for backdoored images, we show evidence that these additional steps correspond to PGD having to reverse the internal backdoor feature, providing a more detailed understanding of the internal mechanisms driving the differences.

\paragraph{Limitations.}
The main limitation of this work is that the analyses we run using the backdoor direction require full access to the trigger. This limits the applicability of a large part of this work to controlled research settings, as practical defenders cannot assume to know the trigger a priori. While the weight orthogonalization removes the backdoor with minimal impact on clean accuracy, using this in practice would require reverse engineering the trigger directly.

Furthermore, while this assumption is not necessary for the weight detection in Section~\ref{sec:detecting}, our method fails to detect several established attacks. While this limitation is strong, we note that detecting the stealthier attacks that are often harder to defend against for ViT defenses~\cite{doan2023defending,subramanya2024closer}, which focus on anomalies in attention mechanisms, is still valuable. Additionally, the scheme does not require any (clean) data and runs in approximately a minute per model. Finally, adaptive attackers with full control over training could relatively straightforwardly bypass the proposed detection scheme (see, e.g.,~\cite{grond,gong2025megatron}), although this is a strong additional attack assumption that is difficult to defend against in practice.

\section{Conclusion and Future Work}
\label{sec:conclusions}

In this work, we explore the use of mechanistic interpretability to understand how varying backdoor attacks are represented inside ViTs. Using knowledge of the trigger mechanism, we first identify linear directions corresponding to the backdoor feature, and subsequently evaluate it by using it to remove the backdoor from the model. Furthermore, we analyze how adversarial examples interact with the backdoor direction. Finally, we introduce a simple diagnostic tool to analyze ViT weights directly to detect backdoored models.

\paragraph{Future Work.} In future work exploring automated techniques to find the direction without knowledge of trigger could be interesting to introduce more realistic defenses. Furthermore, if we assume trigger knowledge, it could be interesting to use automated interpretability tools to derive backdoor circuits (see, .e.g.,~\cite{conmy2023towards}).

%
%
\bibliographystyle{splncs04}
\bibliography{main}

\input{sec/sup}

\end{document}

%% file: sec/sup.tex
\onecolumn
\appendix

\begin{center}
    {\Large \bf Appendix}
\end{center}


\section{Additional Steering Results}
\begin{figure*}[h]
    \centering
    \includegraphics[width=0.8\linewidth]{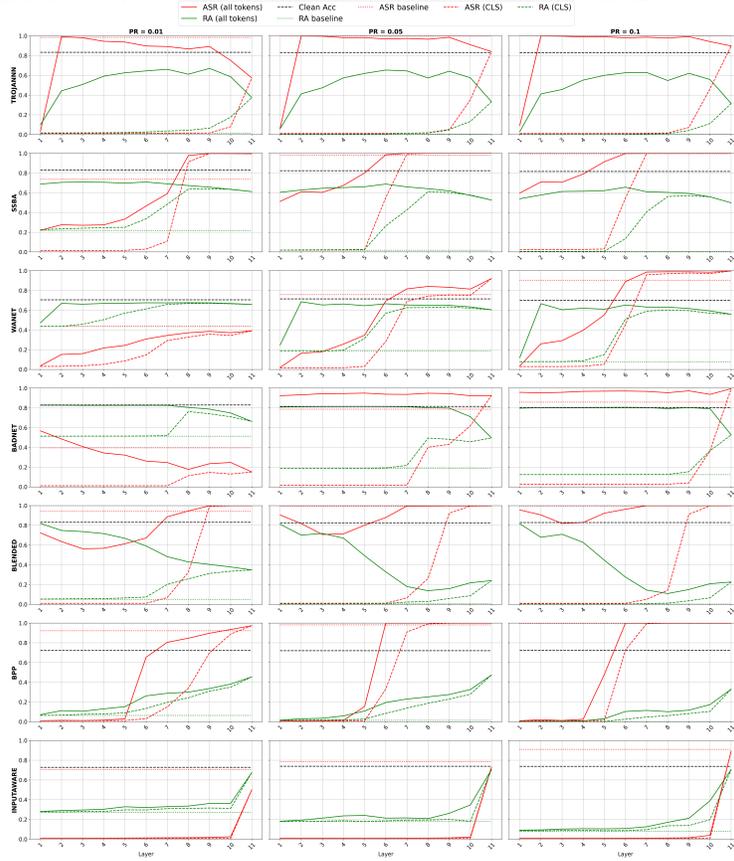}
    \caption{Steering results on CIFAR-100 over seven attacks and three poisoning rates.}
    \label{fig:steering_cifar100}
\end{figure*}

\begin{figure}
    \centering
    \includegraphics[width=0.8\linewidth]{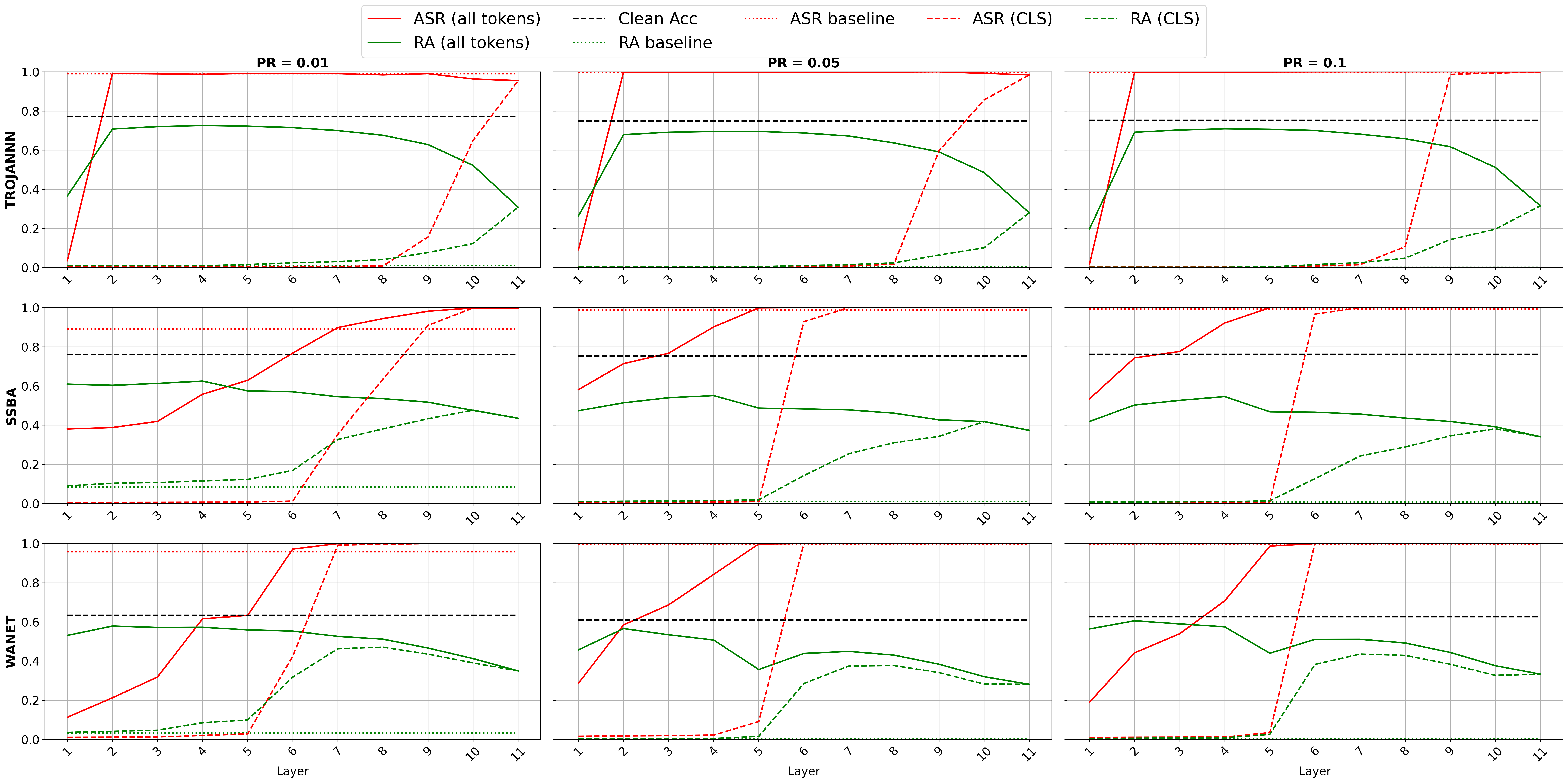}
    \caption{Steering results on Tiny-ImageNet over three attacks and three poisoning rates.}
    \label{fig:tiny_steering}
\end{figure}

\begin{figure}
    \centering
    \includegraphics[width=0.8\linewidth]{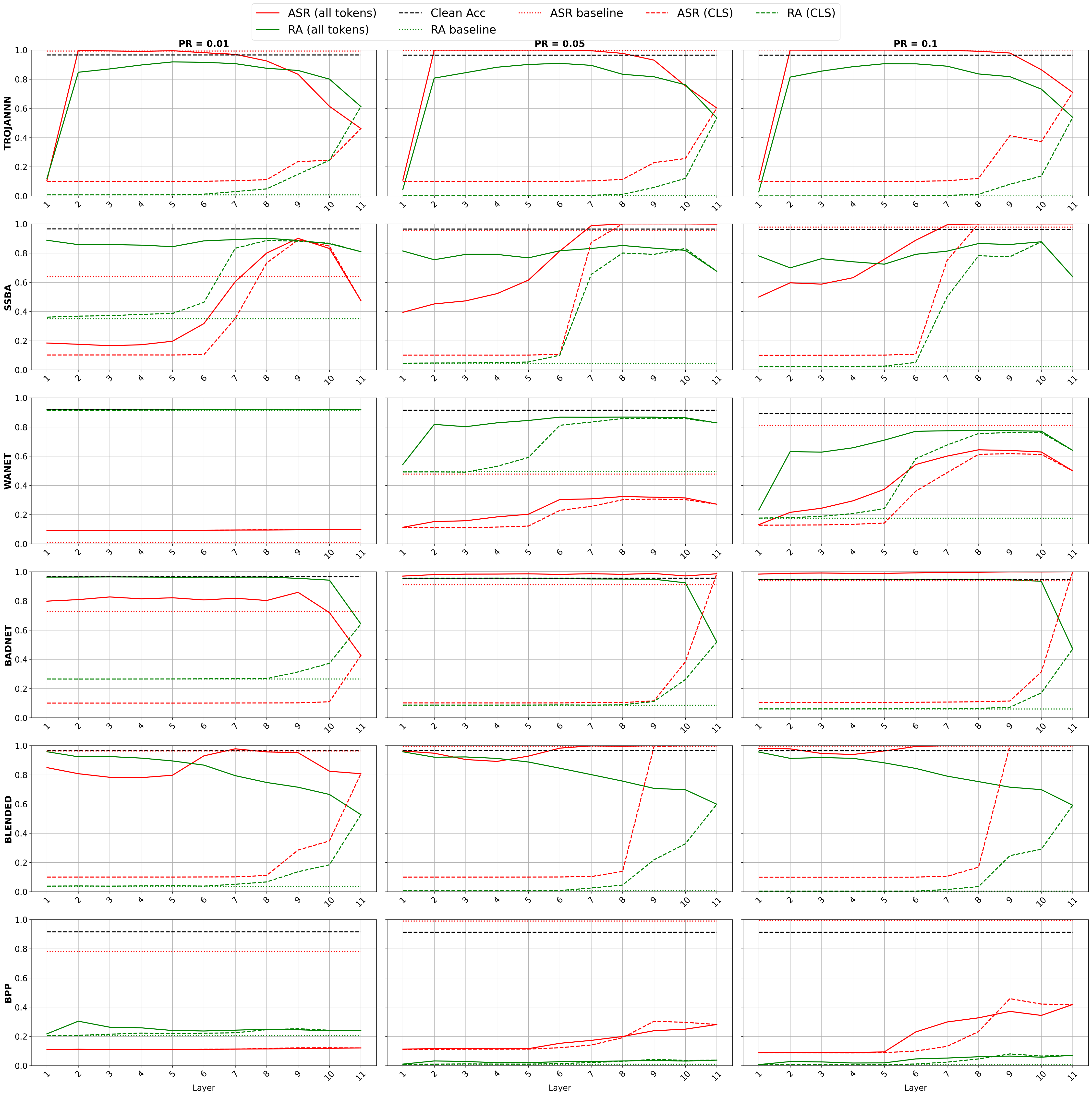}
    \caption{Steering results on CIFAR-10 over six attacks and three poisoning rates.}
    \label{fig:cifar10}
\end{figure}

In this section, we present the steering of the backdoor results across all attacks and poisoning rates for the three considered datasets. Figures~\ref{fig:steering_cifar100}, \ref{fig:tiny_steering} and \ref{fig:cifar10} present these results for CIFAR-100, Tiny-ImageNet and CIFAR-10 datasets, respectively. The number of attacks varies across datasets, as some models were not available for download on BackdoorBench or had low ASR.

\clearpage
\section{Additional Weight Orthogonalization Results}
Table~\ref{tab:ortho_tiny} shows weight orthogonalization results for the Tiny-ImageNet dataset, while Table~\ref{tab:ortho_cifar10} shows these results for CIFAR-10.
In only a few cases, the ASR after weight orthogonalization does not drop below 1\%, while CA always remains close to the baseline results. 
\begin{table*}[t]
\centering
\caption{Weight orthogonalization results on Tiny-ImageNet across poisoning rates (PRs).}
\label{tab:ortho_tiny}
\adjustbox{max width=\textwidth}{
\begin{tabular}{lc|ccccc|ccccc|ccccc|}
\toprule
\multicolumn{2}{c|}{} & \multicolumn{5}{c|}{TROJANNN} & \multicolumn{5}{c|}{SSBA} & \multicolumn{5}{c|}{WANET} \\
 & PR & ASR & RA & CA & Prec & Rec & ASR & RA & CA & Prec & Rec & ASR & RA & CA & Prec & Rec \\
\midrule
\multirow{3}{*}{\textbf{Baseline}} & 0.01 &  99.0 &   1.0 &  77.1 &  92.3 &  96.0 &  89.1 &   8.5 &  76.0 &  85.2 &  92.0 &  95.8 &   3.3 &  63.4 &  49.5 &  94.0 \\
& 0.05 &  99.7 &   0.2 &  74.9 &  81.4 &  96.0 &  98.9 &   0.9 &  75.2 &  81.0 &  94.0 &  99.7 &   0.2 &  60.9 &  28.9 &  86.0 \\
& 0.1 &  99.8 &   0.1 &  75.2 &  86.8 &  92.0 &  99.3 &   0.6 &  76.2 &  87.0 &  94.0 &  99.5 &   0.4 &  62.7 &  49.0 &  94.0 \\
\midrule
\multirow{3}{*}{\textbf{After Orth.}} & 0.01 &   0.0 &  58.4 &  76.8 &  95.7 &  90.0 &   0.0 &  61.4 &  75.8 &  97.9 &  92.0 &   0.1 &  59.1 &  63.1 &  73.3 &  44.0 \\
& 0.05 &   8.0 &  51.9 &  74.8 &  95.9 &  94.0 &   0.0 &  59.7 &  75.0 &  95.7 &  90.0 &   0.8 &  59.4 &  60.6 &  58.2 &  64.0 \\
& 0.1 &   0.8 &  43.7 &  74.9 &  91.7 &  88.0 &   0.1 &  58.5 &  76.0 &  97.8 &  90.0 &   0.7 &  60.4 &  61.6 &  79.5 &  70.0 \\
\bottomrule
\end{tabular}
}
\end{table*}

\begin{table*}[t]
\centering
\caption{Weight orthogonalization results on CIFAR-100 across different poisoning rates (PR) and attacks.}
\label{tab:ortho_cifar100}
\adjustbox{max width=\textwidth}{
\begin{tabular}{lc|ccccc|ccccc|ccccc|ccccc|ccccc|ccccc|}
\toprule
\multicolumn{2}{c|}{} & \multicolumn{5}{c|}{BadNet} & \multicolumn{5}{c|}{Blended} & \multicolumn{5}{c|}{WaNet} & \multicolumn{5}{c|}{SSBA} & \multicolumn{5}{c|}{BPP} & \multicolumn{5}{c|}{TrojanNN} \\
 & PR & ASR & RA & CA & Prec & Rec & ASR & RA & CA & Prec & Rec & ASR & RA & CA & Prec & Rec & ASR & RA & CA & Prec & Rec & ASR & RA & CA & Prec & Rec & ASR & RA & CA & Prec & Rec \\
\midrule
\multirow{3}{*}{\textbf{Baseline}} & 0.01 &  39.5 &  51.3 &  82.9 &  84.3 &  91.0 &  94.1 &   5.0 &  83.1 &  93.7 &  89.0 &  43.5 &  43.7 &  70.5 &  25.8 &  80.0 &  73.6 &  21.8 &  82.8 &  69.2 &  90.0 &  92.2 &   6.3 &  72.1 &  84.0 &  84.0 &  98.4 &   1.5 &  83.3 &  90.3 &  93.0 \\
& 0.05 &  78.7 &  18.7 &  81.5 &  57.6 &  91.0 &  99.5 &   0.5 &  82.5 &  90.2 &  92.0 &  75.8 &  18.8 &  71.4 &  47.5 &  75.0 &  98.0 &   1.7 &  82.3 &  49.7 &  89.0 &  98.2 &   1.5 &  71.9 &  94.8 &  73.0 &  99.7 &   0.3 &  83.1 &  85.7 &  90.0 \\
& 0.1 &  85.8 &  12.4 &  80.2 &  32.7 &  89.0 &  99.7 &   0.2 &  82.8 &  91.8 &  90.0 &  90.4 &   7.7 &  70.0 &  27.8 &  78.0 &  99.5 &   0.4 &  81.6 &  41.4 &  89.0 &  99.7 &   0.2 &  72.2 &  86.6 &  84.0 &  99.9 &   0.1 &  82.8 &  86.4 &  89.0 \\
\midrule
\multirow{3}{*}{\textbf{After Orth.}} & 0.01 &   0.3 &  81.0 &  81.3 &  89.9 &  89.0 &   0.0 &  39.8 &  82.7 &  98.6 &  73.0 &  47.9 &  40.5 &  69.3 &  20.8 &  81.0 &   0.0 &  73.7 &  82.9 & 100.0 &  69.0 &   0.0 &  62.6 &  72.1 &  96.2 &  50.0 &   0.0 &  59.8 &  81.3 &   0.0 &   0.0 \\
& 0.05 &   0.0 &  80.7 &  81.8 &  97.2 &  70.0 &  53.9 &  22.8 &  81.9 &  89.0 &  81.0 &   0.0 &  69.8 &  70.9 &  93.3 &  14.0 &   0.1 &  73.6 &  82.8 & 100.0 &  74.0 &   0.0 &  64.0 &  71.7 & 100.0 &  38.0 &   0.0 &  55.7 &  81.6 &   0.0 &   0.0 \\
& 0.1 &   0.0 &  79.6 &  80.5 & 100.0 &  22.0 &  67.6 &  18.6 &  81.8 &  72.4 &  89.0 &   0.0 &  69.5 &  70.3 & 100.0 &   9.0 &   0.1 &  71.9 &  82.3 &  97.1 &  68.0 &   0.0 &  62.7 &  71.9 & 100.0 &  57.0 &   0.0 &  54.7 &  81.5 &   0.0 &   0.0 \\
\bottomrule
\end{tabular}}
\end{table*}
\begin{table*}
\centering
\caption{Weight orthogonalization results on CIFAR-10 across poisoning rates (PRs).}
\label{tab:ortho_cifar10}
\adjustbox{max width=\textwidth}{
\begin{tabular}{lc|ccccc|ccccc|ccccc|ccccc|ccccc|ccccc|}
\toprule
\multicolumn{2}{c|}{} & \multicolumn{5}{c|}{BADNET} & \multicolumn{5}{c|}{BLENDED} & \multicolumn{5}{c|}{WANET} & \multicolumn{5}{c|}{SSBA} & \multicolumn{5}{c|}{BPP} & \multicolumn{5}{c|}{TROJANNN} \\
 & PR & ASR & RA & CA & Prec & Rec & ASR & RA & CA & Prec & Rec & ASR & RA & CA & Prec & Rec & ASR & RA & CA & Prec & Rec & ASR & RA & CA & Prec & Rec & ASR & RA & CA & Prec & Rec \\
\midrule
\multirow{3}{*}{\textbf{Baseline}} & 0.01 &  72.7 &  26.5 &  96.5 &  96.3 &  96.8 &  96.3 &   3.5 &  96.5 &  97.0 &  96.9 &   0.7 &  91.5 &  92.1 &  96.3 &  87.4 &  63.8 &  35.0 &  96.5 &  95.9 &  98.0 &  78.0 &  20.4 &  91.6 &  87.4 &  96.1 &  99.2 &   0.8 &  96.6 &  97.3 &  97.5 \\
& 0.05 &  91.1 &   8.6 &  95.6 &  94.8 &  96.1 &  99.3 &   0.7 &  96.7 &  97.7 &  97.7 &  47.8 &  49.4 &  91.5 &  83.8 &  92.8 &  95.5 &   4.4 &  96.5 &  95.8 &  97.4 &  99.0 &   1.0 &  91.3 &  86.2 &  96.6 &  99.9 &   0.1 &  96.5 &  97.3 &  97.2 \\
& 0.1 &  93.8 &   6.0 &  94.8 &  90.8 &  95.4 &  99.7 &   0.3 &  96.5 &  97.8 &  97.1 &  81.0 &  17.6 &  89.1 &  74.0 &  94.1 &  97.8 &   2.1 &  96.2 &  96.6 &  96.7 &  99.4 &   0.6 &  91.4 &  96.6 &  84.9 & 100.0 &   0.0 &  96.5 &  97.5 &  97.0 \\
\midrule
\multirow{3}{*}{\textbf{After Orth.}} & 0.01 &   7.6 &  89.1 &  95.6 &  94.9 &  95.2 &   0.1 &  81.8 &  95.8 &  98.9 &  88.7 &   1.6 &  90.4 &  91.7 &  95.5 &  88.3 &   0.6 &  94.0 &  96.5 &  98.1 &  95.8 &  76.2 &  22.1 &  90.8 &  85.5 &  96.5 &   0.0 &  82.6 &  92.0 & 100.0 &  50.4 \\
& 0.05 &   0.4 &  92.5 &  95.3 &  96.5 &  93.2 &   0.0 &  77.7 &  96.3 &  98.6 &  93.8 &  47.8 &  49.2 &  90.9 &  82.0 &  93.1 &   0.5 &  93.7 &  96.3 &  99.1 &  94.1 &   0.9 &  83.7 &  91.0 &  88.3 &  93.2 &   0.0 &  79.8 &  91.5 & 100.0 &  47.5 \\
& 0.1 &   0.8 &  90.7 &  94.9 &  94.7 &  93.9 &   0.0 &  77.6 &  95.9 &  98.8 &  89.9 &   0.2 &  91.0 &  90.5 &  96.7 &  81.8 &   0.5 &  93.3 &  95.9 &  99.1 &  91.8 &   0.2 &  82.2 &  90.5 &  96.8 &  76.5 &   0.0 &  79.5 &  92.0 & 100.0 &  51.3 \\
\bottomrule
\end{tabular}
}
\end{table*}

\clearpage
\section{Class Distributions AEs}
\label{app:distributions}
Here we present the resulting class distributions of adversarial examples. In each Figure, the top row is the PGD starting from clean images, with $0$ as the target class. In the bottom row, we begin with backdoored images. Note that since the label for the backdoored image is always $0$, no images get misclassified as class $0$. We make a special case for the original label (before poisoning) and set it to $0$ class.
\begin{figure*}
    \centering
    \includegraphics[width=0.8\linewidth]{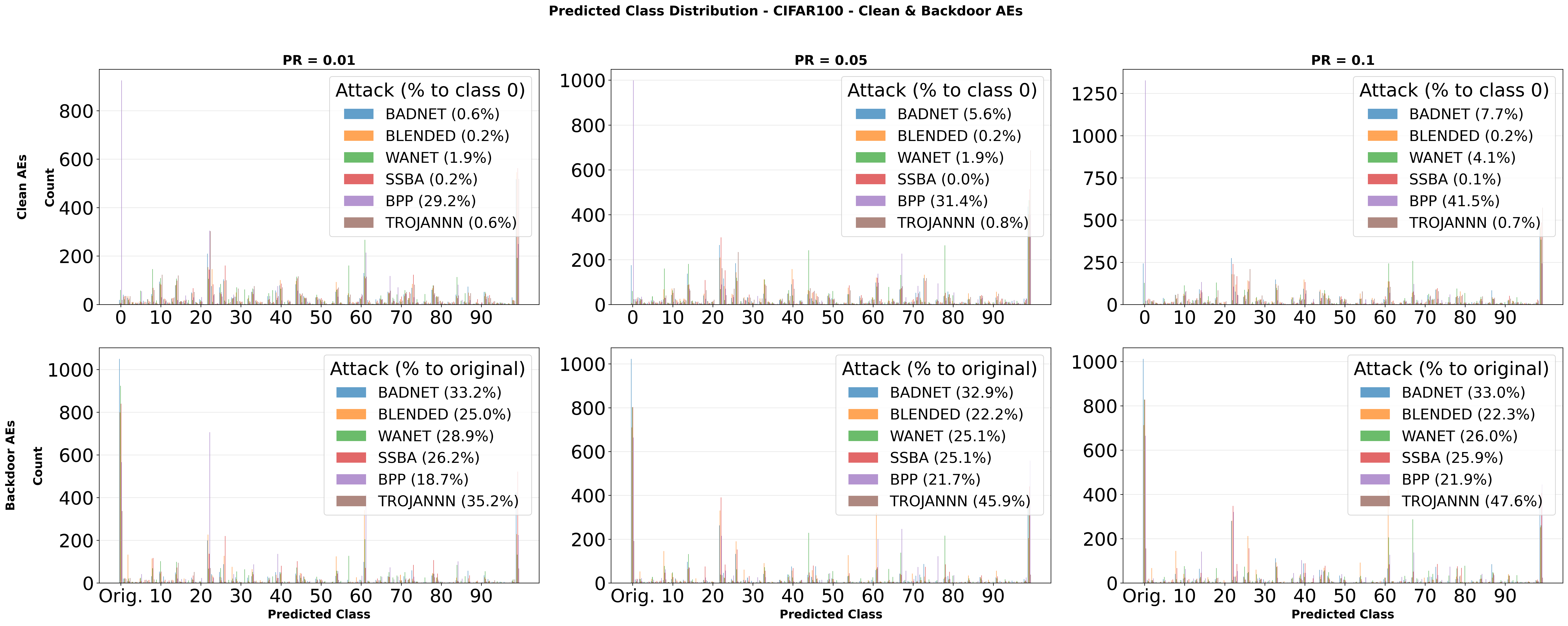}
    \caption{Distributions of predicted classes for backdoored models on CIFAR-100. Starting from clean images (above) and poisoned images (below).}
    \label{fig:cifar100_distributions}
\end{figure*}
\begin{figure*}
    \centering
    \includegraphics[width=0.8\linewidth]{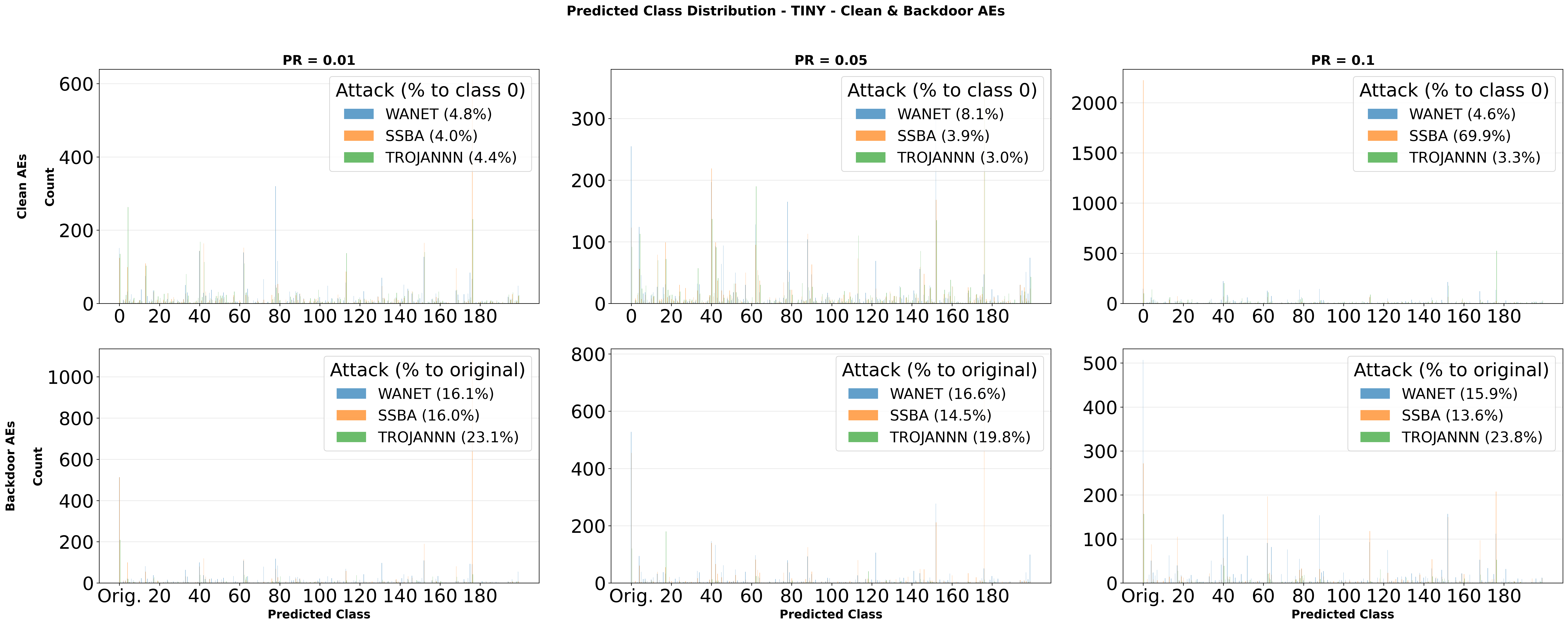}
    \caption{Distributions of predicted classes for backdoored models on Tiny-ImageNet. Starting from clean images (above) and poisoned images (below).}
    \label{fig:tiny_distributions}
\end{figure*}
\begin{figure*}
    \centering
    \includegraphics[width=0.8\linewidth]{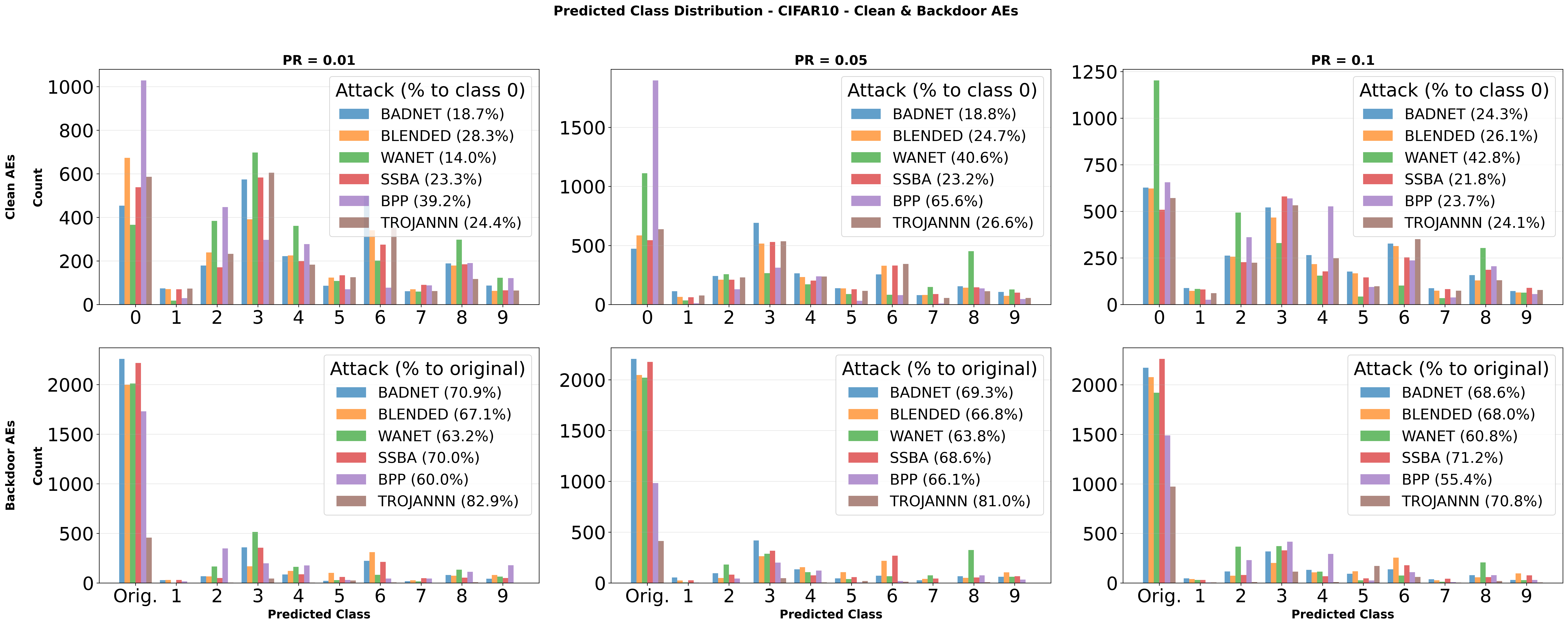}
    \caption{Distributions of predicted classes for backdoored models on CIFAR-10. Starting from clean images (above) and poisoned images (below).}
    \label{fig:cifar10_distributions}
\end{figure*}

\clearpage
\section{Additional Cosine Similarity Tables for Starting From Clean Images}
Cosine similarity for all models is shown in this section. References for specific dataset/poisoning rate settings are in Table~\ref{tab:table_for_tables}. The dataset/poisoning rate for each table is also indicated in the respective caption.
\begin{table}
    \centering
\begin{tabular}{l|c|c|c}
         \toprule
         Poisoning Rate& 0.01 &0.05  &0.1\\
         \midrule
         CIFAR10 Rate& \ref{tab:cosine_cifar10_001} &\ref{tab:cosine_cifar10_005}&\ref{tab:cosine_cifar10_01}\\
         CIFAR100 Rate& \ref{tab:cosine_cifar100_001} &\ref{tab:cosine_cifar100_0_05_app}&\ref{tab:cosine_cifar100_01}\\
         Tiny-ImageNet Rate& \ref{tab:cosine_tiny_001} &\ref{tab:cosine_tiny_005}&\ref{tab:cosine_tiny_01}\\
    \end{tabular}
    \caption{Tables for each setting, for AEs starting from clean images}
    \label{tab:table_for_tables}
\end{table}

\begin{table*}[htbp]
\centering
\caption{Mean Cosine Similarity CIFAR-100 0.1: Target (Tgt) vs. Non-Target (Oth) Class (Median)}
\label{tab:cosine_cifar100_01}
\adjustbox{max width=\linewidth}{
\begin{tabular}{llcccccccccccccc}
\toprule
PR & Layer & \multicolumn{2}{c}{BADNET} & \multicolumn{2}{c}{BLENDED} & \multicolumn{2}{c}{WANET} & \multicolumn{2}{c}{INPUTAWARE} & \multicolumn{2}{c}{SSBA} & \multicolumn{2}{c}{BPP} & \multicolumn{2}{c}{TROJANNN} \\
 &  & Tgt & Oth & Tgt & Oth & Tgt & Oth & Tgt & Oth & Tgt & Oth & Tgt & Oth & Tgt & Oth \\
\midrule
0.1 & $N$ & 243 & 2957 & 7 & 3193 & 128 & 3072 & 0 & 3200 & 2 & 3198 & 1327 & 1873 & 22 & 3178 \\
 & L1 & 0.08$\pm$0.18 (0.11) & 0.06$\pm$0.17 (0.08) & 0.35$\pm$0.20 (0.28) & 0.51$\pm$0.23 (0.55) & 0.11$\pm$0.18 (0.11) & 0.25$\pm$0.16 (0.26) & - & -0.47$\pm$0.15 (-0.49) & -0.54$\pm$0.08 (-0.54) & -0.59$\pm$0.12 (-0.61) & -0.41$\pm$0.13 (-0.42) & -0.32$\pm$0.17 (-0.33) & -0.29$\pm$0.12 (-0.30) & -0.34$\pm$0.15 (-0.35) \\
 & L2 & 0.02$\pm$0.11 (0.05) & 0.01$\pm$0.10 (0.03) & 0.26$\pm$0.20 (0.18) & 0.33$\pm$0.17 (0.35) & 0.10$\pm$0.18 (0.12) & 0.26$\pm$0.15 (0.27) & - & -0.46$\pm$0.12 (-0.47) & -0.56$\pm$0.08 (-0.56) & -0.60$\pm$0.13 (-0.63) & -0.38$\pm$0.12 (-0.38) & -0.28$\pm$0.15 (-0.29) & -0.29$\pm$0.09 (-0.29) & -0.29$\pm$0.14 (-0.30) \\
 & L3 & -0.08$\pm$0.11 (-0.06) & -0.10$\pm$0.11 (-0.09) & 0.22$\pm$0.13 (0.15) & 0.15$\pm$0.15 (0.15) & -0.02$\pm$0.25 (0.01) & -0.14$\pm$0.18 (-0.13) & - & -0.32$\pm$0.12 (-0.34) & -0.46$\pm$0.06 (-0.46) & -0.54$\pm$0.11 (-0.56) & 0.21$\pm$0.13 (0.22) & 0.17$\pm$0.11 (0.18) & -0.18$\pm$0.09 (-0.18) & -0.11$\pm$0.12 (-0.10) \\
 & L4 & 0.01$\pm$0.09 (0.03) & -0.01$\pm$0.09 (0.00) & 0.12$\pm$0.15 (0.08) & 0.12$\pm$0.13 (0.12) & -0.11$\pm$0.21 (-0.10) & -0.21$\pm$0.15 (-0.21) & - & -0.21$\pm$0.10 (-0.22) & -0.30$\pm$0.03 (-0.30) & -0.35$\pm$0.11 (-0.36) & 0.23$\pm$0.10 (0.23) & 0.18$\pm$0.08 (0.18) & -0.00$\pm$0.09 (0.02) & 0.09$\pm$0.10 (0.10) \\
\bottomrule
\end{tabular}}
\end{table*}

\begin{table*}[htbp]
\centering
\caption{Mean Cosine Similarity CIFAR-100 0.05: Target (Tgt) vs. Non-Target (Oth) Class (Median)}
\label{tab:cosine_cifar100_0_05_app}
\adjustbox{max width=\linewidth}{
\begin{tabular}{llcccccccccccccc}
\toprule
PR & Layer & \multicolumn{2}{c}{BADNET} & \multicolumn{2}{c}{BLENDED} & \multicolumn{2}{c}{WANET} & \multicolumn{2}{c}{INPUTAWARE} & \multicolumn{2}{c}{SSBA} & \multicolumn{2}{c}{BPP} & \multicolumn{2}{c}{TROJANNN} \\
 &  & Tgt & Oth & Tgt & Oth & Tgt & Oth & Tgt & Oth & Tgt & Oth & Tgt & Oth & Tgt & Oth \\
\midrule
0.05 & $N$ & 175 & 3025 & 7 & 3193 & 60 & 3140 & 10 & 3190 & 1 & 3199 & 999 & 2201 & 24 & 3176 \\
 & L1 & 0.05$\pm$0.21 (0.08) & 0.02$\pm$0.17 (0.04) & 0.43$\pm$0.26 (0.30) & 0.54$\pm$0.23 (0.59) & 0.23$\pm$0.14 (0.25) & 0.29$\pm$0.13 (0.30) & -0.34$\pm$0.16 (-0.34) & -0.33$\pm$0.15 (-0.34) & -0.59$\pm$0.00 (-0.59) & -0.57$\pm$0.12 (-0.59) & -0.45$\pm$0.14 (-0.45) & -0.38$\pm$0.16 (-0.38) & -0.28$\pm$0.17 (-0.28) & -0.32$\pm$0.17 (-0.33) \\
 & L2 & -0.03$\pm$0.13 (0.00) & -0.04$\pm$0.11 (-0.02) & 0.31$\pm$0.22 (0.19) & 0.35$\pm$0.18 (0.37) & 0.23$\pm$0.18 (0.23) & 0.32$\pm$0.14 (0.33) & -0.42$\pm$0.14 (-0.44) & -0.38$\pm$0.12 (-0.40) & -0.65$\pm$0.00 (-0.65) & -0.60$\pm$0.12 (-0.62) & -0.45$\pm$0.10 (-0.45) & -0.39$\pm$0.12 (-0.40) & -0.30$\pm$0.11 (-0.27) & -0.30$\pm$0.13 (-0.31) \\
 & L3 & -0.08$\pm$0.13 (-0.05) & -0.10$\pm$0.11 (-0.08) & 0.26$\pm$0.14 (0.17) & 0.17$\pm$0.15 (0.16) & -0.21$\pm$0.17 (-0.17) & -0.21$\pm$0.15 (-0.21) & -0.32$\pm$0.14 (-0.37) & -0.21$\pm$0.14 (-0.24) & -0.55$\pm$0.00 (-0.55) & -0.54$\pm$0.11 (-0.56) & 0.08$\pm$0.12 (0.07) & 0.04$\pm$0.12 (0.03) & -0.23$\pm$0.09 (-0.22) & -0.14$\pm$0.12 (-0.13) \\
 & L4 & 0.01$\pm$0.10 (0.02) & -0.00$\pm$0.09 (0.00) & 0.14$\pm$0.14 (0.08) & 0.14$\pm$0.13 (0.15) & -0.20$\pm$0.15 (-0.20) & -0.21$\pm$0.13 (-0.20) & -0.13$\pm$0.10 (-0.12) & -0.04$\pm$0.10 (-0.04) & -0.38$\pm$0.00 (-0.38) & -0.37$\pm$0.10 (-0.39) & 0.11$\pm$0.10 (0.11) & 0.07$\pm$0.09 (0.07) & 0.01$\pm$0.08 (0.01) & 0.08$\pm$0.10 (0.09) \\
\bottomrule
\end{tabular}}
\end{table*}

\begin{table*}[htbp]
\centering
\caption{Mean Cosine Similarity CIFAR-100 0.01: Target (Tgt) vs. Non-Target (Oth) Class(Median)}
\label{tab:cosine_cifar100_001}
\adjustbox{max width=\linewidth}{
\begin{tabular}{llcccccccccccccc}
\toprule
PR & Layer & \multicolumn{2}{c}{BADNET} & \multicolumn{2}{c}{BLENDED} & \multicolumn{2}{c}{WANET} & \multicolumn{2}{c}{INPUTAWARE} & \multicolumn{2}{c}{SSBA} & \multicolumn{2}{c}{BPP} & \multicolumn{2}{c}{TROJANNN} \\
 &  & Tgt & Oth & Tgt & Oth & Tgt & Oth & Tgt & Oth & Tgt & Oth & Tgt & Oth & Tgt & Oth \\
\midrule
0.01 & $N$ & 19 & 3181 & 7 & 3193 & 60 & 3140 & 16 & 3184 & 6 & 3194 & 925 & 2275 & 19 & 3181 \\
 & L1 & -0.01$\pm$0.21 (0.05) & 0.01$\pm$0.18 (0.05) & 0.51$\pm$0.24 (0.68) & 0.55$\pm$0.23 (0.60) & 0.35$\pm$0.14 (0.36) & 0.43$\pm$0.15 (0.43) & -0.26$\pm$0.21 (-0.31) & -0.25$\pm$0.23 (-0.28) & -0.60$\pm$0.04 (-0.60) & -0.63$\pm$0.12 (-0.65) & -0.49$\pm$0.13 (-0.49) & -0.47$\pm$0.15 (-0.49) & -0.32$\pm$0.13 (-0.31) & -0.34$\pm$0.15 (-0.35) \\
 & L2 & -0.02$\pm$0.14 (-0.02) & -0.03$\pm$0.10 (-0.01) & 0.38$\pm$0.21 (0.51) & 0.32$\pm$0.19 (0.34) & 0.37$\pm$0.16 (0.39) & 0.51$\pm$0.13 (0.52) & -0.46$\pm$0.11 (-0.46) & -0.38$\pm$0.14 (-0.39) & -0.56$\pm$0.04 (-0.54) & -0.61$\pm$0.13 (-0.64) & -0.49$\pm$0.10 (-0.50) & -0.45$\pm$0.12 (-0.46) & -0.32$\pm$0.09 (-0.33) & -0.29$\pm$0.13 (-0.30) \\
 & L3 & -0.08$\pm$0.16 (-0.06) & -0.04$\pm$0.12 (-0.03) & 0.27$\pm$0.16 (0.36) & 0.16$\pm$0.15 (0.16) & 0.07$\pm$0.15 (0.09) & 0.10$\pm$0.12 (0.11) & -0.37$\pm$0.08 (-0.40) & -0.26$\pm$0.13 (-0.28) & -0.43$\pm$0.09 (-0.45) & -0.54$\pm$0.11 (-0.56) & 0.15$\pm$0.12 (0.15) & 0.08$\pm$0.13 (0.08) & -0.19$\pm$0.08 (-0.19) & -0.11$\pm$0.12 (-0.10) \\
 & L4 & 0.01$\pm$0.12 (0.00) & 0.05$\pm$0.10 (0.06) & 0.24$\pm$0.17 (0.32) & 0.16$\pm$0.13 (0.17) & -0.03$\pm$0.16 (0.00) & -0.06$\pm$0.12 (-0.06) & -0.24$\pm$0.08 (-0.28) & -0.15$\pm$0.11 (-0.17) & -0.32$\pm$0.11 (-0.36) & -0.42$\pm$0.10 (-0.43) & 0.07$\pm$0.09 (0.07) & 0.04$\pm$0.09 (0.04) & 0.00$\pm$0.08 (0.01) & 0.08$\pm$0.10 (0.09) \\
\bottomrule
\end{tabular}}
\end{table*}

\begin{table*}[htbp]
\centering
\caption{Mean Cosine Similarity CIFAR-10 0.1: Target (Tgt) vs. Non-Target (Oth) Class (Median)}
\label{tab:cosine_cifar10_01}
\adjustbox{max width=\linewidth}{
\begin{tabular}{llcccccccccccc}
\toprule
PR & Layer & \multicolumn{2}{c}{BADNET} & \multicolumn{2}{c}{BLENDED} & \multicolumn{2}{c}{WANET} & \multicolumn{2}{c}{SSBA} & \multicolumn{2}{c}{BPP} & \multicolumn{2}{c}{TROJANNN} \\
 &  & Tgt & Oth & Tgt & Oth & Tgt & Oth & Tgt & Oth & Tgt & Oth & Tgt & Oth \\
\midrule
0.1 & $N$ & 628 & 2572 & 623 & 2577 & 1203 & 1997 & 509 & 2691 & 656 & 2544 & 571 & 2629 \\
 & L1 & -0.00$\pm$0.15 (-0.02) & -0.01$\pm$0.13 (-0.01) & 0.54$\pm$0.18 (0.54) & 0.58$\pm$0.18 (0.60) & 0.44$\pm$0.16 (0.45) & 0.47$\pm$0.15 (0.48) & -0.64$\pm$0.13 (-0.66) & -0.66$\pm$0.13 (-0.68) & -0.38$\pm$0.11 (-0.39) & -0.34$\pm$0.13 (-0.35) & -0.44$\pm$0.14 (-0.46) & -0.37$\pm$0.16 (-0.39) \\
 & L2 & -0.05$\pm$0.10 (-0.05) & -0.06$\pm$0.09 (-0.05) & 0.44$\pm$0.16 (0.45) & 0.48$\pm$0.16 (0.50) & 0.49$\pm$0.12 (0.49) & 0.50$\pm$0.12 (0.50) & -0.66$\pm$0.10 (-0.68) & -0.66$\pm$0.11 (-0.68) & -0.41$\pm$0.13 (-0.42) & -0.34$\pm$0.15 (-0.35) & -0.45$\pm$0.11 (-0.47) & -0.41$\pm$0.13 (-0.43) \\
 & L3 & -0.15$\pm$0.09 (-0.16) & -0.16$\pm$0.08 (-0.16) & 0.31$\pm$0.11 (0.31) & 0.34$\pm$0.12 (0.35) & 0.21$\pm$0.14 (0.21) & 0.18$\pm$0.14 (0.19) & -0.61$\pm$0.09 (-0.63) & -0.60$\pm$0.09 (-0.62) & -0.10$\pm$0.09 (-0.10) & -0.07$\pm$0.10 (-0.07) & -0.34$\pm$0.11 (-0.34) & -0.31$\pm$0.12 (-0.32) \\
 & L4 & -0.00$\pm$0.09 (-0.00) & -0.02$\pm$0.10 (-0.02) & 0.19$\pm$0.10 (0.20) & 0.22$\pm$0.11 (0.22) & -0.04$\pm$0.16 (-0.05) & -0.08$\pm$0.16 (-0.08) & -0.48$\pm$0.08 (-0.50) & -0.48$\pm$0.08 (-0.49) & -0.02$\pm$0.10 (-0.03) & 0.01$\pm$0.10 (0.01) & -0.09$\pm$0.10 (-0.09) & -0.07$\pm$0.11 (-0.07) \\
 & L5 & -0.08$\pm$0.06 (-0.08) & -0.08$\pm$0.06 (-0.08) & 0.14$\pm$0.10 (0.14) & 0.17$\pm$0.10 (0.17) & -0.09$\pm$0.16 (-0.10) & -0.15$\pm$0.15 (-0.15) & -0.48$\pm$0.08 (-0.49) & -0.48$\pm$0.09 (-0.49) & 0.02$\pm$0.10 (0.03) & 0.04$\pm$0.10 (0.05) & -0.06$\pm$0.08 (-0.06) & -0.05$\pm$0.09 (-0.04) \\
\bottomrule
\end{tabular}
}
\end{table*}

\begin{table*}[htbp]
\centering
\caption{Mean Cosine Similarity CIFAR-10 0.05: Target (Tgt) vs. Non-Target (Oth) Class (Median)}
\label{tab:cosine_cifar10_005}
\adjustbox{max width=\linewidth}{
\begin{tabular}{llcccccccccccc}
\toprule
PR & Layer & \multicolumn{2}{c}{BADNET} & \multicolumn{2}{c}{BLENDED} & \multicolumn{2}{c}{WANET} & \multicolumn{2}{c}{SSBA} & \multicolumn{2}{c}{BPP} & \multicolumn{2}{c}{TROJANNN} \\
 &  & Tgt & Oth & Tgt & Oth & Tgt & Oth & Tgt & Oth & Tgt & Oth & Tgt & Oth \\
\midrule
0.05 & $N$ & 472 & 2728 & 585 & 2615 & 1111 & 2089 & 544 & 2656 & 1898 & 1302 & 637 & 2563 \\
 & L1 & -0.03$\pm$0.14 (-0.03) & -0.01$\pm$0.13 (-0.01) & 0.54$\pm$0.18 (0.54) & 0.58$\pm$0.19 (0.61) & 0.46$\pm$0.13 (0.46) & 0.49$\pm$0.12 (0.49) & -0.67$\pm$0.11 (-0.69) & -0.67$\pm$0.12 (-0.70) & -0.45$\pm$0.12 (-0.46) & -0.41$\pm$0.14 (-0.42) & -0.46$\pm$0.14 (-0.48) & -0.41$\pm$0.15 (-0.43) \\
 & L2 & -0.12$\pm$0.10 (-0.12) & -0.09$\pm$0.10 (-0.10) & 0.41$\pm$0.15 (0.41) & 0.45$\pm$0.16 (0.47) & 0.44$\pm$0.12 (0.44) & 0.46$\pm$0.12 (0.47) & -0.69$\pm$0.10 (-0.71) & -0.69$\pm$0.10 (-0.71) & -0.46$\pm$0.14 (-0.48) & -0.41$\pm$0.15 (-0.42) & -0.43$\pm$0.11 (-0.45) & -0.39$\pm$0.13 (-0.41) \\
 & L3 & -0.18$\pm$0.08 (-0.17) & -0.17$\pm$0.08 (-0.17) & 0.26$\pm$0.11 (0.27) & 0.29$\pm$0.12 (0.30) & 0.30$\pm$0.13 (0.31) & 0.29$\pm$0.13 (0.30) & -0.63$\pm$0.09 (-0.65) & -0.62$\pm$0.09 (-0.64) & -0.11$\pm$0.11 (-0.13) & -0.11$\pm$0.11 (-0.12) & -0.32$\pm$0.11 (-0.33) & -0.29$\pm$0.12 (-0.30) \\
 & L4 & 0.00$\pm$0.09 (0.01) & -0.00$\pm$0.10 (0.00) & 0.21$\pm$0.11 (0.22) & 0.24$\pm$0.12 (0.24) & 0.10$\pm$0.17 (0.10) & 0.08$\pm$0.17 (0.09) & -0.49$\pm$0.08 (-0.50) & -0.48$\pm$0.09 (-0.49) & -0.04$\pm$0.08 (-0.04) & -0.02$\pm$0.09 (-0.02) & -0.12$\pm$0.10 (-0.12) & -0.10$\pm$0.11 (-0.10) \\
 & L5 & -0.10$\pm$0.06 (-0.09) & -0.09$\pm$0.07 (-0.09) & 0.10$\pm$0.10 (0.11) & 0.12$\pm$0.10 (0.12) & 0.08$\pm$0.18 (0.09) & 0.03$\pm$0.17 (0.04) & -0.50$\pm$0.08 (-0.51) & -0.50$\pm$0.08 (-0.51) & 0.04$\pm$0.09 (0.03) & 0.03$\pm$0.09 (0.03) & -0.09$\pm$0.08 (-0.09) & -0.08$\pm$0.09 (-0.08) \\
\bottomrule
\end{tabular}
}
\end{table*}

\begin{table*}[htbp]
\centering
\caption{Mean Cosine Similarity CIFAR-10 0.01: Target (Tgt) vs. Non-Target (Oth) Class(Median)}
\label{tab:cosine_cifar10_001}
\adjustbox{max width=\linewidth}{
\begin{tabular}{llcccccccccccc}
\toprule
PR & Layer & \multicolumn{2}{c}{BADNET} & \multicolumn{2}{c}{BLENDED} & \multicolumn{2}{c}{WANET} & \multicolumn{2}{c}{SSBA} & \multicolumn{2}{c}{BPP} & \multicolumn{2}{c}{TROJANNN} \\
 &  & Tgt & Oth & Tgt & Oth & Tgt & Oth & Tgt & Oth & Tgt & Oth & Tgt & Oth \\
\midrule
0.01 & $N$ & 454 & 2746 & 673 & 2527 & 366 & 2834 & 538 & 2662 & 1029 & 2171 & 586 & 2614 \\
 & L1 & -0.03$\pm$0.12 (-0.03) & -0.05$\pm$0.12 (-0.04) & 0.53$\pm$0.18 (0.54) & 0.57$\pm$0.20 (0.60) & 0.35$\pm$0.16 (0.35) & 0.39$\pm$0.15 (0.40) & -0.68$\pm$0.12 (-0.70) & -0.69$\pm$0.12 (-0.72) & -0.51$\pm$0.11 (-0.52) & -0.48$\pm$0.13 (-0.50) & -0.45$\pm$0.14 (-0.47) & -0.37$\pm$0.16 (-0.39) \\
 & L2 & -0.17$\pm$0.10 (-0.17) & -0.17$\pm$0.09 (-0.17) & 0.42$\pm$0.16 (0.44) & 0.47$\pm$0.17 (0.49) & 0.44$\pm$0.14 (0.44) & 0.45$\pm$0.14 (0.46) & -0.72$\pm$0.10 (-0.74) & -0.72$\pm$0.11 (-0.75) & -0.47$\pm$0.12 (-0.48) & -0.43$\pm$0.13 (-0.44) & -0.43$\pm$0.10 (-0.45) & -0.37$\pm$0.13 (-0.39) \\
 & L3 & -0.13$\pm$0.09 (-0.13) & -0.15$\pm$0.09 (-0.14) & 0.30$\pm$0.12 (0.31) & 0.33$\pm$0.13 (0.34) & 0.42$\pm$0.13 (0.44) & 0.44$\pm$0.13 (0.45) & -0.68$\pm$0.08 (-0.70) & -0.66$\pm$0.10 (-0.68) & -0.19$\pm$0.11 (-0.20) & -0.16$\pm$0.12 (-0.17) & -0.32$\pm$0.10 (-0.33) & -0.28$\pm$0.12 (-0.29) \\
 & L4 & 0.00$\pm$0.09 (0.01) & -0.01$\pm$0.09 (-0.00) & 0.22$\pm$0.11 (0.22) & 0.25$\pm$0.12 (0.26) & 0.13$\pm$0.16 (0.14) & 0.14$\pm$0.16 (0.14) & -0.53$\pm$0.08 (-0.54) & -0.52$\pm$0.08 (-0.53) & -0.05$\pm$0.09 (-0.05) & -0.02$\pm$0.10 (-0.02) & -0.08$\pm$0.10 (-0.07) & -0.06$\pm$0.11 (-0.06) \\
 & L5 & -0.10$\pm$0.06 (-0.10) & -0.10$\pm$0.07 (-0.09) & 0.22$\pm$0.10 (0.22) & 0.24$\pm$0.10 (0.24) & 0.10$\pm$0.14 (0.10) & 0.10$\pm$0.15 (0.09) & -0.50$\pm$0.08 (-0.51) & -0.50$\pm$0.08 (-0.51) & 0.03$\pm$0.10 (0.04) & 0.06$\pm$0.10 (0.06) & 0.01$\pm$0.08 (0.02) & 0.02$\pm$0.09 (0.02) \\
\bottomrule
\end{tabular}
}
\end{table*}

\begin{table*}[htbp]
\centering
\caption{Mean Cosine Similarity Tiny-ImageNet 0.1: Target (Tgt) vs. Non-Target (Oth) Class (Median)}
\label{tab:cosine_tiny_01}
\adjustbox{max width=\linewidth}{
\begin{tabular}{llcccccc}
\toprule
PR & Layer & \multicolumn{2}{c}{WANET} & \multicolumn{2}{c}{SSBA} & \multicolumn{2}{c}{TROJANNN} \\
 &  & Tgt & Oth & Tgt & Oth & Tgt & Oth \\
\midrule
0.1 & $N$ & 146 & 3054 & 2224 & 976 & 100 & 3100 \\
 & L1 & 0.28$\pm$0.14 (0.26) & 0.41$\pm$0.13 (0.41) & -0.41$\pm$0.16 (-0.42) & -0.34$\pm$0.19 (-0.35) & -0.10$\pm$0.23 (-0.07) & -0.20$\pm$0.22 (-0.20) \\
 & L2 & 0.27$\pm$0.16 (0.27) & 0.40$\pm$0.12 (0.42) & -0.40$\pm$0.13 (-0.41) & -0.32$\pm$0.15 (-0.31) & -0.24$\pm$0.16 (-0.23) & -0.28$\pm$0.16 (-0.29) \\
 & L3 & -0.01$\pm$0.17 (0.02) & 0.06$\pm$0.13 (0.07) & -0.19$\pm$0.11 (-0.19) & -0.15$\pm$0.13 (-0.15) & -0.08$\pm$0.14 (-0.07) & -0.11$\pm$0.13 (-0.11) \\
 & L4 & 0.04$\pm$0.17 (0.07) & 0.07$\pm$0.12 (0.08) & 0.00$\pm$0.09 (0.00) & -0.04$\pm$0.11 (-0.04) & 0.01$\pm$0.11 (0.01) & -0.01$\pm$0.10 (-0.01) \\
 & L5 & 0.21$\pm$0.22 (0.24) & 0.01$\pm$0.17 (0.01) & 0.42$\pm$0.10 (0.43) & 0.21$\pm$0.16 (0.21) & 0.07$\pm$0.07 (0.08) & 0.06$\pm$0.06 (0.07) \\
 & L6 & 0.63$\pm$0.33 (0.77) & 0.07$\pm$0.17 (0.07) & 0.81$\pm$0.09 (0.83) & 0.25$\pm$0.31 (0.13) & 0.06$\pm$0.05 (0.06) & 0.05$\pm$0.05 (0.06) \\
 & L7 & 0.59$\pm$0.36 (0.74) & 0.02$\pm$0.12 (0.01) & 0.80$\pm$0.08 (0.82) & 0.27$\pm$0.30 (0.15) & 0.06$\pm$0.05 (0.07) & 0.05$\pm$0.06 (0.06) \\
 & L8 & 0.56$\pm$0.36 (0.71) & -0.00$\pm$0.12 (-0.02) & 0.80$\pm$0.08 (0.82) & 0.28$\pm$0.28 (0.17) & 0.05$\pm$0.06 (0.06) & 0.03$\pm$0.06 (0.03) \\
 & L9 & 0.57$\pm$0.36 (0.74) & 0.01$\pm$0.12 (-0.00) & 0.85$\pm$0.07 (0.86) & 0.25$\pm$0.33 (0.09) & 0.04$\pm$0.06 (0.04) & -0.00$\pm$0.06 (-0.00) \\
 & L10 & 0.60$\pm$0.26 (0.69) & 0.01$\pm$0.12 (0.01) & 0.89$\pm$0.05 (0.90) & 0.24$\pm$0.37 (0.07) & 0.22$\pm$0.10 (0.22) & 0.01$\pm$0.07 (0.01) \\
 & L11 & 0.62$\pm$0.32 (0.74) & -0.01$\pm$0.15 (-0.01) & 0.87$\pm$0.06 (0.89) & 0.22$\pm$0.38 (0.07) & 0.23$\pm$0.15 (0.23) & -0.01$\pm$0.08 (-0.01) \\
\bottomrule
\end{tabular}
}
\end{table*}

\begin{table*}[htbp]
\centering
\caption{Mean Cosine Similarity Tiny-ImageNet 0.05: Target (Tgt) vs. Non-Target (Oth) Class (Median)}
\label{tab:cosine_tiny_005}
\adjustbox{max width=\linewidth}{
\begin{tabular}{llcccccc}
\toprule
PR & Layer & \multicolumn{2}{c}{WANET} & \multicolumn{2}{c}{SSBA} & \multicolumn{2}{c}{TROJANNN} \\
 &  & Tgt & Oth & Tgt & Oth & Tgt & Oth \\
\midrule
0.05 & $N$ & 255 & 2945 & 123 & 3077 & 92 & 3108 \\
 & L1 & 0.28$\pm$0.14 (0.29) & 0.43$\pm$0.13 (0.43) & -0.35$\pm$0.17 (-0.36) & -0.36$\pm$0.19 (-0.37) & -0.04$\pm$0.24 (-0.04) & -0.12$\pm$0.23 (-0.12) \\
 & L2 & 0.29$\pm$0.14 (0.29) & 0.43$\pm$0.11 (0.44) & -0.40$\pm$0.13 (-0.42) & -0.39$\pm$0.14 (-0.39) & -0.20$\pm$0.15 (-0.21) & -0.24$\pm$0.16 (-0.24) \\
 & L3 & -0.04$\pm$0.16 (-0.02) & 0.04$\pm$0.12 (0.05) & -0.18$\pm$0.13 (-0.18) & -0.18$\pm$0.13 (-0.18) & -0.04$\pm$0.12 (-0.04) & -0.07$\pm$0.13 (-0.06) \\
 & L4 & 0.03$\pm$0.16 (0.06) & 0.05$\pm$0.12 (0.06) & -0.08$\pm$0.10 (-0.09) & -0.10$\pm$0.10 (-0.10) & 0.07$\pm$0.09 (0.07) & 0.05$\pm$0.10 (0.05) \\
 & L5 & 0.30$\pm$0.22 (0.30) & 0.06$\pm$0.17 (0.06) & 0.10$\pm$0.10 (0.11) & 0.04$\pm$0.10 (0.03) & 0.09$\pm$0.07 (0.10) & 0.08$\pm$0.07 (0.08) \\
 & L6 & 0.70$\pm$0.29 (0.82) & 0.05$\pm$0.15 (0.05) & 0.30$\pm$0.25 (0.27) & 0.04$\pm$0.09 (0.04) & 0.11$\pm$0.06 (0.12) & 0.10$\pm$0.06 (0.10) \\
 & L7 & 0.69$\pm$0.30 (0.81) & 0.00$\pm$0.13 (-0.01) & 0.32$\pm$0.27 (0.34) & 0.06$\pm$0.09 (0.05) & 0.10$\pm$0.06 (0.10) & 0.08$\pm$0.07 (0.09) \\
 & L8 & 0.66$\pm$0.30 (0.78) & 0.00$\pm$0.12 (-0.00) & 0.36$\pm$0.25 (0.37) & 0.10$\pm$0.08 (0.10) & 0.06$\pm$0.06 (0.07) & 0.03$\pm$0.07 (0.03) \\
 & L9 & 0.70$\pm$0.30 (0.82) & 0.01$\pm$0.12 (-0.00) & 0.33$\pm$0.27 (0.31) & 0.03$\pm$0.08 (0.03) & 0.04$\pm$0.07 (0.05) & -0.01$\pm$0.07 (-0.01) \\
 & L10 & 0.68$\pm$0.24 (0.77) & 0.01$\pm$0.13 (-0.00) & 0.55$\pm$0.26 (0.55) & 0.04$\pm$0.11 (0.03) & 0.22$\pm$0.10 (0.23) & 0.03$\pm$0.07 (0.03) \\
 & L11 & 0.72$\pm$0.29 (0.85) & -0.01$\pm$0.16 (-0.01) & 0.52$\pm$0.28 (0.54) & 0.02$\pm$0.12 (0.02) & 0.25$\pm$0.16 (0.27) & -0.00$\pm$0.08 (0.00) \\
\bottomrule
\end{tabular}
}
\end{table*}

\begin{table*}[htbp]
\centering
\caption{Mean Cosine Similarity Tiny-ImageNet 0.01: Target (Tgt) vs. Non-Target (Oth) Class (Median)}
\label{tab:cosine_tiny_001}
\adjustbox{max width=\linewidth}{
\begin{tabular}{llcccccc}
\toprule
PR & Layer & \multicolumn{2}{c}{WANET} & \multicolumn{2}{c}{SSBA} & \multicolumn{2}{c}{TROJANNN} \\
 &  & Tgt & Oth & Tgt & Oth & Tgt & Oth \\
\midrule
0.01 & $N$ & 151 & 3049 & 124 & 3076 & 135 & 3065 \\
 & L1 & 0.27$\pm$0.17 (0.27) & 0.38$\pm$0.17 (0.40) & -0.36$\pm$0.17 (-0.38) & -0.44$\pm$0.18 (-0.45) & -0.13$\pm$0.20 (-0.12) & -0.23$\pm$0.21 (-0.23) \\
 & L2 & 0.32$\pm$0.16 (0.31) & 0.43$\pm$0.14 (0.43) & -0.36$\pm$0.12 (-0.36) & -0.40$\pm$0.14 (-0.40) & -0.29$\pm$0.12 (-0.29) & -0.32$\pm$0.15 (-0.33) \\
 & L3 & 0.03$\pm$0.17 (0.03) & 0.11$\pm$0.15 (0.12) & -0.20$\pm$0.12 (-0.20) & -0.24$\pm$0.12 (-0.24) & -0.11$\pm$0.09 (-0.10) & -0.11$\pm$0.11 (-0.11) \\
 & L4 & 0.08$\pm$0.17 (0.08) & 0.02$\pm$0.15 (0.03) & -0.12$\pm$0.10 (-0.12) & -0.14$\pm$0.10 (-0.15) & -0.01$\pm$0.07 (-0.01) & -0.02$\pm$0.10 (-0.02) \\
 & L5 & 0.12$\pm$0.20 (0.09) & 0.05$\pm$0.16 (0.06) & -0.05$\pm$0.10 (-0.06) & -0.05$\pm$0.09 (-0.05) & 0.06$\pm$0.05 (0.06) & 0.05$\pm$0.06 (0.06) \\
 & L6 & 0.20$\pm$0.32 (0.14) & 0.01$\pm$0.16 (0.02) & -0.07$\pm$0.08 (-0.07) & -0.09$\pm$0.08 (-0.09) & 0.07$\pm$0.05 (0.08) & 0.06$\pm$0.05 (0.06) \\
 & L7 & 0.23$\pm$0.39 (0.06) & -0.07$\pm$0.13 (-0.08) & -0.01$\pm$0.08 (-0.02) & -0.02$\pm$0.07 (-0.02) & 0.06$\pm$0.06 (0.07) & 0.05$\pm$0.06 (0.05) \\
 & L8 & 0.23$\pm$0.37 (0.06) & -0.06$\pm$0.11 (-0.06) & 0.01$\pm$0.07 (0.02) & 0.00$\pm$0.07 (0.00) & 0.05$\pm$0.06 (0.06) & 0.02$\pm$0.06 (0.02) \\
 & L9 & 0.28$\pm$0.38 (0.05) & -0.04$\pm$0.09 (-0.04) & -0.02$\pm$0.06 (-0.02) & -0.03$\pm$0.07 (-0.03) & 0.04$\pm$0.07 (0.04) & 0.01$\pm$0.06 (0.01) \\
 & L10 & 0.46$\pm$0.22 (0.42) & -0.01$\pm$0.10 (-0.01) & 0.30$\pm$0.12 (0.31) & 0.02$\pm$0.08 (0.02) & 0.17$\pm$0.08 (0.17) & 0.01$\pm$0.06 (0.01) \\
 & L11 & 0.45$\pm$0.31 (0.44) & -0.02$\pm$0.13 (-0.01) & 0.30$\pm$0.17 (0.30) & 0.01$\pm$0.10 (0.01) & 0.21$\pm$0.14 (0.21) & -0.01$\pm$0.08 (-0.01) \\
\bottomrule
\end{tabular}
}
\end{table*}
\newpage
\clearpage
\section{Additional Cosine Similarity Tables for Starting From Backdoored Images}
\label{app:start_bd_ae}
Cosine similarity for all models is shown in this section. References for specific dataset/poisoning rate settings are in Table~\ref{tab:table_for_tables_rev}. The dataset/poisoning rate for each table is also indicated in the respective caption.
\begin{table}[h]
    \centering
    \begin{tabular}{l|c|c|c}
         \toprule
         Poisoning Rate& 0.01 &0.05  &0.1\\
         \midrule
         CIFAR10 Rate& \ref{tab:cosine_cifar10_0_01_rev} &\ref{tab:cosine_cifar10_0_05_rev}&\ref{tab:cosine_cifar10_0_1_rev}\\
         CIFAR100 Rate& \ref{tab:cosine_cifar100_0_01_rev} &\ref{tab:cosine_cifar100_0_05_rev_app}&\ref{tab:cosine_cifar100_0_1_rev}\\
         Tiny-ImageNet Rate& \ref{tab:cosine_tiny_0_01_rev} &\ref{tab:cosine_tiny_0_05_rev}&\ref{tab:cosine_tiny_0_1_rev}\\
    \end{tabular}
    \caption{Tables for each setting, for AEs starting from backdoored images}
    \label{tab:table_for_tables_rev}
\end{table}

\begin{table*}[htbp]
\centering
\caption{Mean Cosine Similarity CIFAR-100 0.01: Original (Org) vs. Other (Oth) Class (Median)}
\label{tab:cosine_cifar100_0_01_rev}
\adjustbox{max width=\linewidth}{\begin{tabular}{llcccccccccccccc}
\toprule
PR & Layer & \multicolumn{2}{c}{BADNET} & \multicolumn{2}{c}{BLENDED} & \multicolumn{2}{c}{WANET} & \multicolumn{2}{c}{INPUTAWARE} & \multicolumn{2}{c}{SSBA} & \multicolumn{2}{c}{BPP} & \multicolumn{2}{c}{TROJANNN} \\
 &  & Tgt & Oth & Tgt & Oth & Tgt & Oth & Tgt & Oth & Tgt & Oth & Tgt & Oth & Tgt & Oth \\
\midrule
0.01 & $N$ & 1050 & 2150 & 800 & 2400 & 924 & 2276 & 620 & 2580 & 840 & 2360 & 566 & 2634 & 337 & 2863 \\
 & L1 & -0.04$\pm$0.19 (0.01) & 0.01$\pm$0.17 (0.04) & 0.42$\pm$0.27 (0.43) & 0.42$\pm$0.26 (0.45) & 0.41$\pm$0.16 (0.42) & 0.39$\pm$0.15 (0.39) & -0.26$\pm$0.23 (-0.29) & -0.27$\pm$0.23 (-0.29) & -0.65$\pm$0.13 (-0.68) & -0.67$\pm$0.11 (-0.69) & -0.56$\pm$0.13 (-0.58) & -0.53$\pm$0.13 (-0.55) & -0.37$\pm$0.15 (-0.38) & -0.37$\pm$0.15 (-0.36) \\
 & L2 & -0.06$\pm$0.11 (-0.04) & -0.04$\pm$0.09 (-0.02) & 0.18$\pm$0.20 (0.16) & 0.14$\pm$0.20 (0.15) & 0.45$\pm$0.16 (0.46) & 0.48$\pm$0.12 (0.49) & -0.35$\pm$0.15 (-0.37) & -0.37$\pm$0.13 (-0.38) & -0.63$\pm$0.13 (-0.66) & -0.67$\pm$0.11 (-0.70) & -0.56$\pm$0.09 (-0.57) & -0.55$\pm$0.09 (-0.57) & -0.30$\pm$0.13 (-0.30) & -0.33$\pm$0.12 (-0.33) \\
 & L3 & -0.10$\pm$0.12 (-0.09) & -0.04$\pm$0.12 (-0.03) & 0.03$\pm$0.15 (0.02) & -0.00$\pm$0.14 (-0.01) & -0.09$\pm$0.16 (-0.08) & -0.05$\pm$0.13 (-0.04) & -0.17$\pm$0.17 (-0.20) & -0.24$\pm$0.13 (-0.26) & -0.57$\pm$0.11 (-0.59) & -0.61$\pm$0.08 (-0.63) & -0.22$\pm$0.11 (-0.23) & -0.23$\pm$0.10 (-0.23) & -0.18$\pm$0.13 (-0.17) & -0.17$\pm$0.12 (-0.16) \\
 & L4 & -0.01$\pm$0.10 (-0.00) & 0.04$\pm$0.09 (0.05) & -0.01$\pm$0.14 (-0.01) & -0.03$\pm$0.12 (-0.03) & -0.26$\pm$0.15 (-0.27) & -0.23$\pm$0.13 (-0.23) & -0.11$\pm$0.13 (-0.13) & -0.15$\pm$0.10 (-0.16) & -0.49$\pm$0.09 (-0.50) & -0.50$\pm$0.08 (-0.51) & -0.23$\pm$0.08 (-0.23) & -0.23$\pm$0.07 (-0.23) & -0.06$\pm$0.11 (-0.06) & -0.02$\pm$0.10 (-0.01) \\
\bottomrule
\end{tabular}}
\end{table*}

\begin{table*}[htbp]
\centering
\caption{Mean Cosine Similarity CIFAR-100 0.05: Original (Org) vs. Other (Oth) Class (Median)}
\label{tab:cosine_cifar100_0_05_rev_app}
\adjustbox{max width=\linewidth}{\begin{tabular}{llcccccccccccccc}
\toprule
PR & Layer & \multicolumn{2}{c}{BADNET} & \multicolumn{2}{c}{BLENDED} & \multicolumn{2}{c}{WANET} & \multicolumn{2}{c}{INPUTAWARE} & \multicolumn{2}{c}{SSBA} & \multicolumn{2}{c}{BPP} & \multicolumn{2}{c}{TROJANNN} \\
 &  & Org & Oth & Org & Oth & Org & Oth & Org & Oth & Org & Oth & Org & Oth & Org & Oth \\
\midrule
0.05 & $N$ & 1023 & 2177 & 710 & 2490 & 802 & 2398 & 676 & 2524 & 803 & 2397 & 664 & 2536 & 192 & 3008 \\
 & L1 & -0.02$\pm$0.17 (0.01) & 0.02$\pm$0.16 (0.04) & 0.41$\pm$0.26 (0.41) & 0.42$\pm$0.25 (0.45) & 0.25$\pm$0.14 (0.26) & 0.25$\pm$0.13 (0.26) & -0.34$\pm$0.15 (-0.36) & -0.35$\pm$0.14 (-0.36) & -0.61$\pm$0.12 (-0.63) & -0.63$\pm$0.11 (-0.65) & -0.49$\pm$0.15 (-0.49) & -0.47$\pm$0.14 (-0.48) & -0.33$\pm$0.17 (-0.35) & -0.34$\pm$0.16 (-0.35) \\
 & L2 & -0.07$\pm$0.11 (-0.05) & -0.05$\pm$0.10 (-0.03) & 0.20$\pm$0.20 (0.19) & 0.18$\pm$0.19 (0.19) & 0.25$\pm$0.16 (0.27) & 0.27$\pm$0.13 (0.28) & -0.37$\pm$0.13 (-0.39) & -0.40$\pm$0.10 (-0.42) & -0.62$\pm$0.12 (-0.64) & -0.66$\pm$0.10 (-0.68) & -0.52$\pm$0.10 (-0.52) & -0.51$\pm$0.09 (-0.52) & -0.28$\pm$0.14 (-0.29) & -0.33$\pm$0.12 (-0.33) \\
 & L3 & -0.15$\pm$0.11 (-0.14) & -0.10$\pm$0.10 (-0.09) & 0.05$\pm$0.16 (0.04) & 0.02$\pm$0.14 (0.01) & -0.35$\pm$0.15 (-0.36) & -0.35$\pm$0.14 (-0.35) & -0.18$\pm$0.15 (-0.21) & -0.23$\pm$0.13 (-0.24) & -0.58$\pm$0.10 (-0.59) & -0.61$\pm$0.08 (-0.63) & -0.29$\pm$0.10 (-0.30) & -0.30$\pm$0.09 (-0.31) & -0.16$\pm$0.13 (-0.15) & -0.18$\pm$0.11 (-0.17) \\
 & L4 & -0.05$\pm$0.09 (-0.04) & -0.01$\pm$0.08 (-0.00) & -0.04$\pm$0.14 (-0.04) & -0.05$\pm$0.12 (-0.04) & -0.39$\pm$0.13 (-0.40) & -0.38$\pm$0.12 (-0.39) & -0.05$\pm$0.10 (-0.05) & -0.06$\pm$0.09 (-0.06) & -0.45$\pm$0.09 (-0.46) & -0.47$\pm$0.08 (-0.47) & -0.23$\pm$0.09 (-0.24) & -0.22$\pm$0.08 (-0.23) & -0.00$\pm$0.10 (0.01) & 0.02$\pm$0.09 (0.03) \\
\bottomrule
\end{tabular} }
\end{table*}
\begin{table*}[htbp]
\centering
\caption{Mean Cosine Similarity CIFAR-100 0.1: Original (Org) vs. Other (Oth) Class (Median)}
\label{tab:cosine_cifar100_0_1_rev}
\adjustbox{max width=\linewidth}{
\begin{tabular}{llcccccccccccccc}
\toprule
PR & Layer & \multicolumn{2}{c}{BADNET} & \multicolumn{2}{c}{BLENDED} & \multicolumn{2}{c}{WANET} & \multicolumn{2}{c}{INPUTAWARE} & \multicolumn{2}{c}{SSBA} & \multicolumn{2}{c}{BPP} & \multicolumn{2}{c}{TROJANNN} \\
 &  & Tgt & Oth & Tgt & Oth & Tgt & Oth & Tgt & Oth & Tgt & Oth & Tgt & Oth & Tgt & Oth \\
\midrule
0.1 & $N$ & 1012 & 2188 & 713 & 2487 & 828 & 2372 & 685 & 2515 & 828 & 2372 & 665 & 2535 & 156 & 3044 \\
 & L1 & 0.02$\pm$0.17 (0.05) & 0.06$\pm$0.16 (0.09) & 0.38$\pm$0.26 (0.37) & 0.39$\pm$0.24 (0.41) & 0.20$\pm$0.18 (0.22) & 0.18$\pm$0.16 (0.19) & -0.51$\pm$0.14 (-0.53) & -0.51$\pm$0.13 (-0.52) & -0.62$\pm$0.13 (-0.65) & -0.64$\pm$0.11 (-0.66) & -0.45$\pm$0.13 (-0.46) & -0.43$\pm$0.13 (-0.44) & -0.36$\pm$0.16 (-0.37) & -0.37$\pm$0.15 (-0.37) \\
 & L2 & -0.02$\pm$0.10 (0.00) & 0.00$\pm$0.09 (0.02) & 0.17$\pm$0.19 (0.15) & 0.15$\pm$0.18 (0.15) & 0.15$\pm$0.18 (0.18) & 0.18$\pm$0.15 (0.20) & -0.45$\pm$0.12 (-0.47) & -0.48$\pm$0.10 (-0.49) & -0.62$\pm$0.13 (-0.65) & -0.66$\pm$0.10 (-0.69) & -0.44$\pm$0.11 (-0.45) & -0.42$\pm$0.11 (-0.43) & -0.26$\pm$0.15 (-0.27) & -0.32$\pm$0.12 (-0.32) \\
 & L3 & -0.15$\pm$0.11 (-0.14) & -0.11$\pm$0.10 (-0.10) & 0.02$\pm$0.16 (0.01) & -0.01$\pm$0.14 (-0.02) & -0.35$\pm$0.19 (-0.37) & -0.32$\pm$0.18 (-0.33) & -0.32$\pm$0.13 (-0.35) & -0.35$\pm$0.10 (-0.37) & -0.58$\pm$0.10 (-0.60) & -0.62$\pm$0.08 (-0.63) & -0.22$\pm$0.11 (-0.22) & -0.20$\pm$0.11 (-0.20) & -0.12$\pm$0.14 (-0.11) & -0.15$\pm$0.12 (-0.14) \\
 & L4 & -0.04$\pm$0.09 (-0.03) & -0.01$\pm$0.08 (0.00) & -0.07$\pm$0.14 (-0.06) & -0.07$\pm$0.12 (-0.07) & -0.43$\pm$0.15 (-0.45) & -0.41$\pm$0.15 (-0.42) & -0.24$\pm$0.10 (-0.25) & -0.26$\pm$0.09 (-0.27) & -0.45$\pm$0.10 (-0.46) & -0.46$\pm$0.08 (-0.46) & -0.13$\pm$0.08 (-0.13) & -0.11$\pm$0.08 (-0.11) & 0.02$\pm$0.11 (0.03) & 0.03$\pm$0.09 (0.04) \\
\bottomrule
\end{tabular}}
\end{table*}

\begin{table*}[htbp]
\centering
\caption{Mean Cosine Similarity CIFAR-10 0.01: Original (Org) vs. Other (Oth) Class (Median)}
\label{tab:cosine_cifar10_0_01_rev}
\adjustbox{max width=\linewidth}{\begin{tabular}{llcccccccccccc}
\toprule
PR & Layer & \multicolumn{2}{c}{BADNET} & \multicolumn{2}{c}{BLENDED} & \multicolumn{2}{c}{WANET} & \multicolumn{2}{c}{SSBA} & \multicolumn{2}{c}{BPP} & \multicolumn{2}{c}{TROJANNN} \\
 &  & Org & Oth & Org & Oth & Org & Oth & Org & Oth & Org & Oth & Org & Oth \\
\midrule
0.01 & $N$ & 2259 & 941 & 1998 & 1202 & 2010 & 1190 & 2218 & 982 & 1730 & 1470 & 457 & 2743 \\
 & L1 & -0.07$\pm$0.10 (-0.06) & -0.05$\pm$0.11 (-0.05) & 0.48$\pm$0.20 (0.49) & 0.46$\pm$0.21 (0.48) & 0.37$\pm$0.15 (0.38) & 0.34$\pm$0.15 (0.34) & -0.75$\pm$0.10 (-0.77) & -0.75$\pm$0.10 (-0.77) & -0.54$\pm$0.11 (-0.55) & -0.53$\pm$0.10 (-0.55) & -0.42$\pm$0.16 (-0.43) & -0.42$\pm$0.15 (-0.44) \\
 & L2 & -0.19$\pm$0.09 (-0.19) & -0.18$\pm$0.08 (-0.18) & 0.35$\pm$0.18 (0.36) & 0.32$\pm$0.19 (0.33) & 0.41$\pm$0.14 (0.41) & 0.39$\pm$0.14 (0.40) & -0.77$\pm$0.09 (-0.79) & -0.77$\pm$0.09 (-0.79) & -0.52$\pm$0.11 (-0.53) & -0.50$\pm$0.10 (-0.50) & -0.41$\pm$0.12 (-0.42) & -0.41$\pm$0.11 (-0.43) \\
 & L3 & -0.17$\pm$0.08 (-0.16) & -0.15$\pm$0.07 (-0.14) & 0.23$\pm$0.13 (0.23) & 0.21$\pm$0.13 (0.21) & 0.35$\pm$0.14 (0.36) & 0.35$\pm$0.13 (0.36) & -0.71$\pm$0.08 (-0.72) & -0.72$\pm$0.08 (-0.74) & -0.38$\pm$0.10 (-0.39) & -0.37$\pm$0.09 (-0.38) & -0.34$\pm$0.11 (-0.34) & -0.33$\pm$0.10 (-0.34) \\
 & L4 & -0.03$\pm$0.08 (-0.03) & 0.00$\pm$0.08 (0.01) & 0.10$\pm$0.13 (0.11) & 0.10$\pm$0.13 (0.10) & 0.00$\pm$0.16 (0.00) & -0.01$\pm$0.15 (-0.02) & -0.58$\pm$0.07 (-0.59) & -0.59$\pm$0.06 (-0.59) & -0.26$\pm$0.09 (-0.27) & -0.26$\pm$0.09 (-0.26) & -0.18$\pm$0.10 (-0.16) & -0.14$\pm$0.10 (-0.13) \\
 & L5 & -0.12$\pm$0.06 (-0.11) & -0.10$\pm$0.05 (-0.10) & 0.08$\pm$0.11 (0.08) & 0.08$\pm$0.11 (0.08) & -0.03$\pm$0.15 (-0.04) & -0.04$\pm$0.13 (-0.05) & -0.58$\pm$0.06 (-0.58) & -0.57$\pm$0.06 (-0.58) & -0.20$\pm$0.09 (-0.19) & -0.18$\pm$0.09 (-0.18) & -0.12$\pm$0.09 (-0.11) & -0.07$\pm$0.07 (-0.06) \\
\bottomrule
\end{tabular}}
\end{table*}
\begin{table*}[htbp]
\centering
\caption{Mean Cosine Similarity CIFAR-10 0.05: Original (Org) vs. Other (Oth) Class (Median)}
\label{tab:cosine_cifar10_0_05_rev}
\adjustbox{max width=\linewidth}{\begin{tabular}{llcccccccccccc}
\toprule
PR & Layer & \multicolumn{2}{c}{BADNET} & \multicolumn{2}{c}{BLENDED} & \multicolumn{2}{c}{WANET} & \multicolumn{2}{c}{SSBA} & \multicolumn{2}{c}{BPP} & \multicolumn{2}{c}{TROJANNN} \\
 &  & Org & Oth & Org & Oth & Org & Oth & Org & Oth & Org & Oth & Org & Oth \\
\midrule
0.05 & $N$ & 2205 & 995 & 2046 & 1154 & 2020 & 1180 & 2176 & 1024 & 983 & 2217 & 414 & 2786 \\
 & L1 & -0.04$\pm$0.11 (-0.03) & -0.02$\pm$0.12 (-0.02) & 0.50$\pm$0.20 (0.50) & 0.48$\pm$0.20 (0.50) & 0.45$\pm$0.13 (0.46) & 0.44$\pm$0.12 (0.45) & -0.73$\pm$0.10 (-0.75) & -0.73$\pm$0.09 (-0.75) & -0.49$\pm$0.11 (-0.50) & -0.50$\pm$0.11 (-0.51) & -0.47$\pm$0.14 (-0.48) & -0.45$\pm$0.14 (-0.47) \\
 & L2 & -0.11$\pm$0.09 (-0.11) & -0.10$\pm$0.09 (-0.10) & 0.33$\pm$0.17 (0.34) & 0.31$\pm$0.17 (0.32) & 0.42$\pm$0.12 (0.43) & 0.41$\pm$0.12 (0.42) & -0.74$\pm$0.09 (-0.76) & -0.74$\pm$0.08 (-0.76) & -0.52$\pm$0.12 (-0.53) & -0.53$\pm$0.11 (-0.54) & -0.43$\pm$0.12 (-0.44) & -0.42$\pm$0.11 (-0.44) \\
 & L3 & -0.19$\pm$0.07 (-0.19) & -0.18$\pm$0.07 (-0.18) & 0.19$\pm$0.13 (0.19) & 0.17$\pm$0.12 (0.18) & 0.15$\pm$0.15 (0.17) & 0.18$\pm$0.14 (0.19) & -0.67$\pm$0.07 (-0.69) & -0.69$\pm$0.07 (-0.70) & -0.40$\pm$0.10 (-0.41) & -0.41$\pm$0.09 (-0.42) & -0.35$\pm$0.11 (-0.35) & -0.33$\pm$0.10 (-0.34) \\
 & L4 & -0.03$\pm$0.09 (-0.03) & 0.00$\pm$0.09 (0.02) & 0.09$\pm$0.13 (0.11) & 0.09$\pm$0.13 (0.10) & -0.12$\pm$0.16 (-0.12) & -0.13$\pm$0.16 (-0.14) & -0.56$\pm$0.07 (-0.57) & -0.55$\pm$0.06 (-0.56) & -0.31$\pm$0.09 (-0.31) & -0.32$\pm$0.07 (-0.32) & -0.18$\pm$0.10 (-0.17) & -0.14$\pm$0.09 (-0.14) \\
 & L5 & -0.12$\pm$0.07 (-0.12) & -0.11$\pm$0.06 (-0.11) & -0.05$\pm$0.10 (-0.05) & -0.05$\pm$0.11 (-0.05) & -0.25$\pm$0.17 (-0.25) & -0.22$\pm$0.16 (-0.24) & -0.61$\pm$0.06 (-0.61) & -0.59$\pm$0.06 (-0.60) & -0.30$\pm$0.09 (-0.30) & -0.29$\pm$0.08 (-0.29) & -0.18$\pm$0.08 (-0.17) & -0.14$\pm$0.07 (-0.13) \\
\bottomrule
\end{tabular}}
\end{table*}
\begin{table*}[htbp]
\centering
\caption{Mean Cosine Similarity CIFAR-10 0.1: Original (Org) vs. Other (Oth) Class (Median)}
\label{tab:cosine_cifar10_0_1_rev}
\adjustbox{max width=\linewidth}{\begin{tabular}{llcccccccccccc}
\toprule
PR & Layer & \multicolumn{2}{c}{BADNET} & \multicolumn{2}{c}{BLENDED} & \multicolumn{2}{c}{WANET} & \multicolumn{2}{c}{SSBA} & \multicolumn{2}{c}{BPP} & \multicolumn{2}{c}{TROJANNN} \\
 &  & Org & Oth & Org & Oth & Org & Oth & Org & Oth & Org & Oth & Org & Oth \\
\midrule
0.1 & $N$ & 2171 & 1029 & 2076 & 1124 & 1919 & 1281 & 2261 & 939 & 1489 & 1711 & 973 & 2227 \\
 & L1 & -0.03$\pm$0.11 (-0.03) & -0.01$\pm$0.13 (-0.01) & 0.49$\pm$0.19 (0.49) & 0.47$\pm$0.19 (0.49) & 0.43$\pm$0.16 (0.44) & 0.41$\pm$0.16 (0.43) & -0.72$\pm$0.11 (-0.74) & -0.72$\pm$0.10 (-0.73) & -0.42$\pm$0.11 (-0.43) & -0.42$\pm$0.10 (-0.43) & -0.42$\pm$0.15 (-0.44) & -0.41$\pm$0.15 (-0.43) \\
 & L2 & -0.08$\pm$0.08 (-0.07) & -0.06$\pm$0.08 (-0.06) & 0.36$\pm$0.17 (0.37) & 0.34$\pm$0.17 (0.36) & 0.44$\pm$0.13 (0.45) & 0.43$\pm$0.12 (0.43) & -0.72$\pm$0.09 (-0.74) & -0.71$\pm$0.09 (-0.73) & -0.44$\pm$0.12 (-0.45) & -0.45$\pm$0.12 (-0.46) & -0.45$\pm$0.11 (-0.46) & -0.45$\pm$0.11 (-0.46) \\
 & L3 & -0.18$\pm$0.07 (-0.18) & -0.16$\pm$0.07 (-0.16) & 0.23$\pm$0.12 (0.24) & 0.21$\pm$0.13 (0.22) & 0.02$\pm$0.15 (0.02) & 0.01$\pm$0.14 (0.02) & -0.66$\pm$0.07 (-0.67) & -0.67$\pm$0.07 (-0.68) & -0.36$\pm$0.09 (-0.37) & -0.36$\pm$0.08 (-0.37) & -0.36$\pm$0.10 (-0.36) & -0.36$\pm$0.10 (-0.36) \\
 & L4 & -0.05$\pm$0.09 (-0.04) & -0.02$\pm$0.09 (-0.01) & 0.07$\pm$0.11 (0.07) & 0.06$\pm$0.11 (0.07) & -0.30$\pm$0.15 (-0.31) & -0.32$\pm$0.13 (-0.33) & -0.57$\pm$0.06 (-0.57) & -0.56$\pm$0.06 (-0.56) & -0.25$\pm$0.09 (-0.25) & -0.25$\pm$0.08 (-0.25) & -0.17$\pm$0.10 (-0.16) & -0.14$\pm$0.10 (-0.14) \\
 & L5 & -0.11$\pm$0.07 (-0.11) & -0.10$\pm$0.06 (-0.10) & -0.02$\pm$0.11 (-0.02) & -0.03$\pm$0.12 (-0.03) & -0.40$\pm$0.14 (-0.41) & -0.39$\pm$0.13 (-0.39) & -0.59$\pm$0.06 (-0.60) & -0.58$\pm$0.06 (-0.59) & -0.29$\pm$0.10 (-0.30) & -0.28$\pm$0.10 (-0.29) & -0.16$\pm$0.08 (-0.15) & -0.12$\pm$0.08 (-0.11) \\
\bottomrule
\end{tabular}}
\end{table*}

\begin{table*}[htbp]
\centering
\caption{Mean Cosine Similarity Tiny-ImageNet 0.01: Original (Org) vs. Other (Oth) Class (Median)}
\label{tab:cosine_tiny_0_01_rev}
\adjustbox{max width=\linewidth}{\begin{tabular}{llcccccc}
\toprule
PR & Layer & \multicolumn{2}{c}{WANET} & \multicolumn{2}{c}{SSBA} & \multicolumn{2}{c}{TROJANNN} \\
 &  & Org & Oth & Org & Oth & Org & Oth \\
\midrule
0.01 & $N$ & 513 & 2687 & 513 & 2687 & 209 & 2991 \\
 & L1 & 0.30$\pm$0.16 (0.31) & 0.26$\pm$0.16 (0.27) & -0.51$\pm$0.17 (-0.54) & -0.56$\pm$0.15 (-0.58) & -0.34$\pm$0.22 (-0.34) & -0.29$\pm$0.19 (-0.29) \\
 & L2 & 0.28$\pm$0.13 (0.27) & 0.28$\pm$0.12 (0.28) & -0.49$\pm$0.13 (-0.50) & -0.53$\pm$0.11 (-0.55) & -0.41$\pm$0.17 (-0.43) & -0.37$\pm$0.13 (-0.38) \\
 & L3 & -0.10$\pm$0.14 (-0.10) & -0.14$\pm$0.13 (-0.14) & -0.38$\pm$0.11 (-0.40) & -0.39$\pm$0.09 (-0.40) & -0.24$\pm$0.12 (-0.23) & -0.17$\pm$0.10 (-0.16) \\
 & L4 & -0.35$\pm$0.18 (-0.35) & -0.34$\pm$0.16 (-0.36) & -0.30$\pm$0.08 (-0.31) & -0.29$\pm$0.07 (-0.30) & -0.13$\pm$0.11 (-0.13) & -0.07$\pm$0.09 (-0.07) \\
 & L5 & -0.44$\pm$0.18 (-0.45) & -0.42$\pm$0.17 (-0.44) & -0.23$\pm$0.08 (-0.23) & -0.21$\pm$0.07 (-0.21) & -0.15$\pm$0.08 (-0.14) & -0.04$\pm$0.09 (-0.03) \\
\bottomrule
\end{tabular}}
\end{table*}
\begin{table*}[htbp]
\centering
\caption{Mean Cosine Similarity Tiny-ImageNet 0.05: Original (Org) vs. Other (Oth) Class (Median)}
\label{tab:cosine_tiny_0_05_rev}
\adjustbox{max width=\linewidth}{\begin{tabular}{llcccccc}
\toprule
PR & Layer & \multicolumn{2}{c}{WANET} & \multicolumn{2}{c}{SSBA} & \multicolumn{2}{c}{TROJANNN} \\
 &  & Org & Oth & Org & Oth & Org & Oth \\
\midrule
0.05 & $N$ & 528 & 2672 & 454 & 2746 & 121 & 3079 \\
 & L1 & 0.30$\pm$0.13 (0.31) & 0.31$\pm$0.11 (0.32) & -0.45$\pm$0.18 (-0.48) & -0.49$\pm$0.15 (-0.51) & -0.21$\pm$0.24 (-0.23) & -0.18$\pm$0.21 (-0.19) \\
 & L2 & 0.26$\pm$0.13 (0.26) & 0.28$\pm$0.11 (0.28) & -0.50$\pm$0.14 (-0.51) & -0.53$\pm$0.11 (-0.55) & -0.29$\pm$0.18 (-0.30) & -0.28$\pm$0.15 (-0.28) \\
 & L3 & -0.16$\pm$0.13 (-0.16) & -0.18$\pm$0.12 (-0.18) & -0.34$\pm$0.11 (-0.36) & -0.36$\pm$0.10 (-0.37) & -0.17$\pm$0.14 (-0.17) & -0.11$\pm$0.11 (-0.10) \\
 & L4 & -0.20$\pm$0.15 (-0.20) & -0.21$\pm$0.14 (-0.22) & -0.28$\pm$0.09 (-0.29) & -0.28$\pm$0.08 (-0.28) & -0.06$\pm$0.11 (-0.05) & -0.00$\pm$0.09 (0.01) \\
 & L5 & -0.53$\pm$0.18 (-0.56) & -0.51$\pm$0.17 (-0.53) & -0.21$\pm$0.09 (-0.21) & -0.19$\pm$0.09 (-0.19) & -0.14$\pm$0.08 (-0.13) & -0.04$\pm$0.09 (-0.03) \\
\bottomrule
\end{tabular}}
\end{table*}
\begin{table*}[htbp]
\centering
\caption{Mean Cosine Similarity Tiny-ImageNet 0.1: Original (Org) vs. Other (Oth) Class (Median)}
\label{tab:cosine_tiny_0_1_rev}
\adjustbox{max width=\linewidth}{\begin{tabular}{llcccccc}
\toprule
PR & Layer & \multicolumn{2}{c}{WANET} & \multicolumn{2}{c}{SSBA} & \multicolumn{2}{c}{TROJANNN} \\
 &  & Org & Oth & Org & Oth & Org & Oth \\
\midrule
0.1 & $N$ & 507 & 2693 & 272 & 2928 & 157 & 3043 \\
 & L1 & 0.28$\pm$0.13 (0.28) & 0.29$\pm$0.11 (0.29) & -0.49$\pm$0.17 (-0.53) & -0.52$\pm$0.14 (-0.54) & -0.28$\pm$0.23 (-0.30) & -0.27$\pm$0.20 (-0.26) \\
 & L2 & 0.22$\pm$0.15 (0.23) & 0.24$\pm$0.13 (0.25) & -0.50$\pm$0.14 (-0.52) & -0.53$\pm$0.11 (-0.54) & -0.36$\pm$0.17 (-0.37) & -0.35$\pm$0.13 (-0.35) \\
 & L3 & -0.13$\pm$0.15 (-0.13) & -0.16$\pm$0.13 (-0.16) & -0.35$\pm$0.10 (-0.37) & -0.35$\pm$0.09 (-0.36) & -0.23$\pm$0.14 (-0.21) & -0.18$\pm$0.11 (-0.17) \\
 & L4 & -0.16$\pm$0.16 (-0.16) & -0.17$\pm$0.14 (-0.17) & -0.23$\pm$0.09 (-0.24) & -0.21$\pm$0.09 (-0.22) & -0.11$\pm$0.12 (-0.10) & -0.08$\pm$0.09 (-0.07) \\
 & L5 & -0.54$\pm$0.19 (-0.56) & -0.51$\pm$0.17 (-0.54) & -0.10$\pm$0.11 (-0.10) & -0.01$\pm$0.13 (-0.02) & -0.20$\pm$0.09 (-0.19) & -0.07$\pm$0.10 (-0.07) \\
\bottomrule
\end{tabular}}
\end{table*}

\clearpage
\section{Backdoor Detection in Weights}



\begin{figure}
    \centering
    \includegraphics[width=0.9\linewidth]{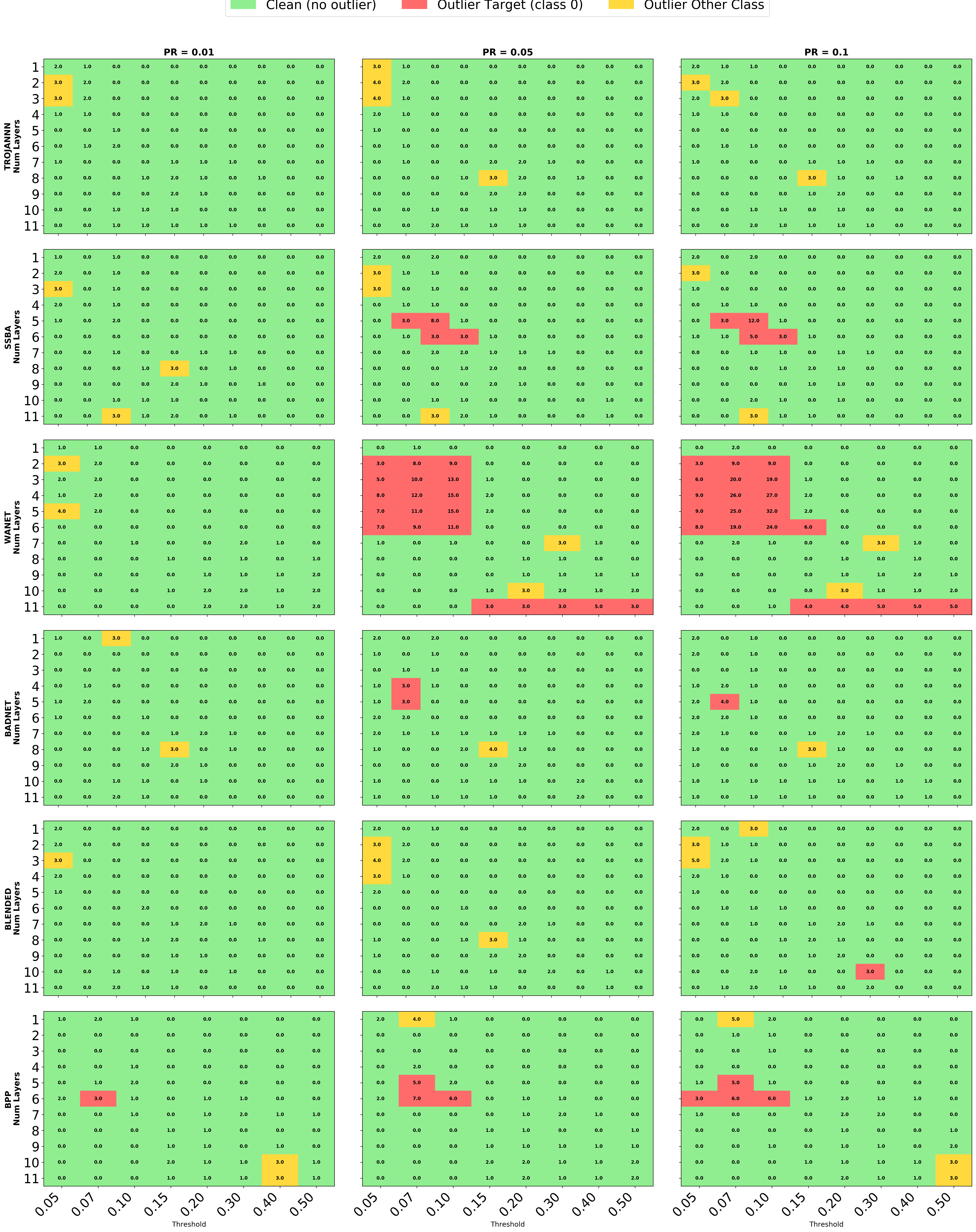}
    \caption{$Z$-scores for CIFAR-10 dataset across different attacks and poisoning rates.}
    \label{fig:cifar10detect}
\end{figure}

\begin{figure}
    \centering
    \includegraphics[width=0.8\linewidth]{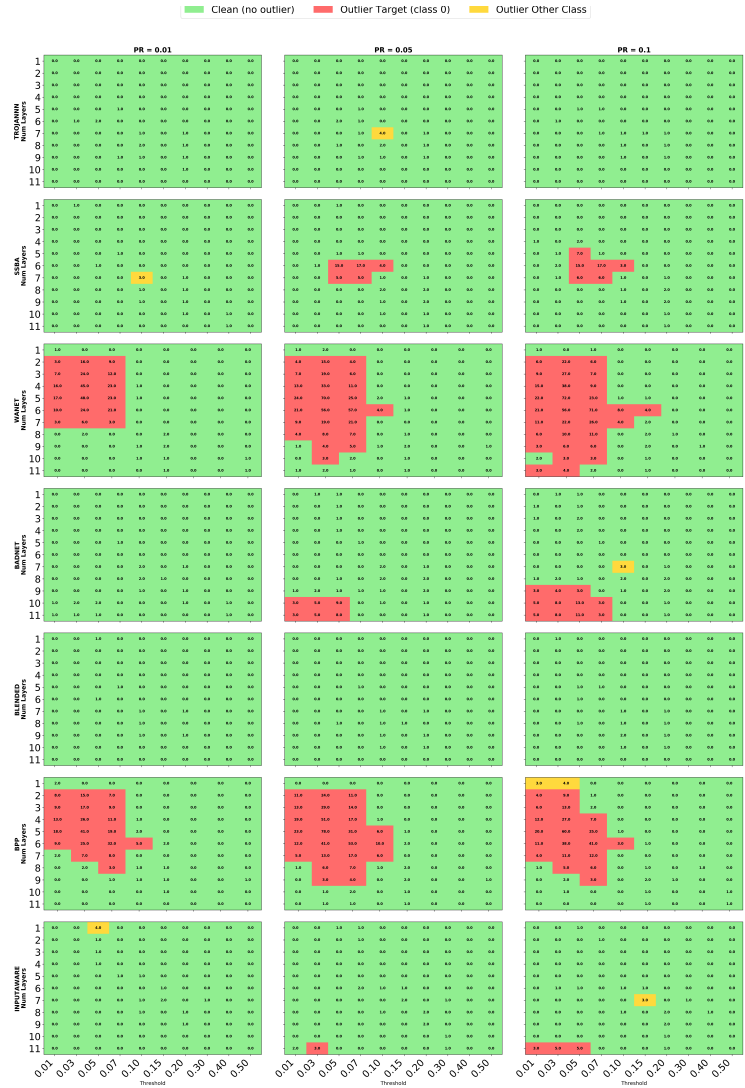}
    \caption{$Z$-scores for CIFAR-100 dataset across different attacks and poisoning rates.}
    \label{fig:cifar100detect}
\end{figure}

\begin{figure}
    \centering
    \includegraphics[width=0.9\linewidth]{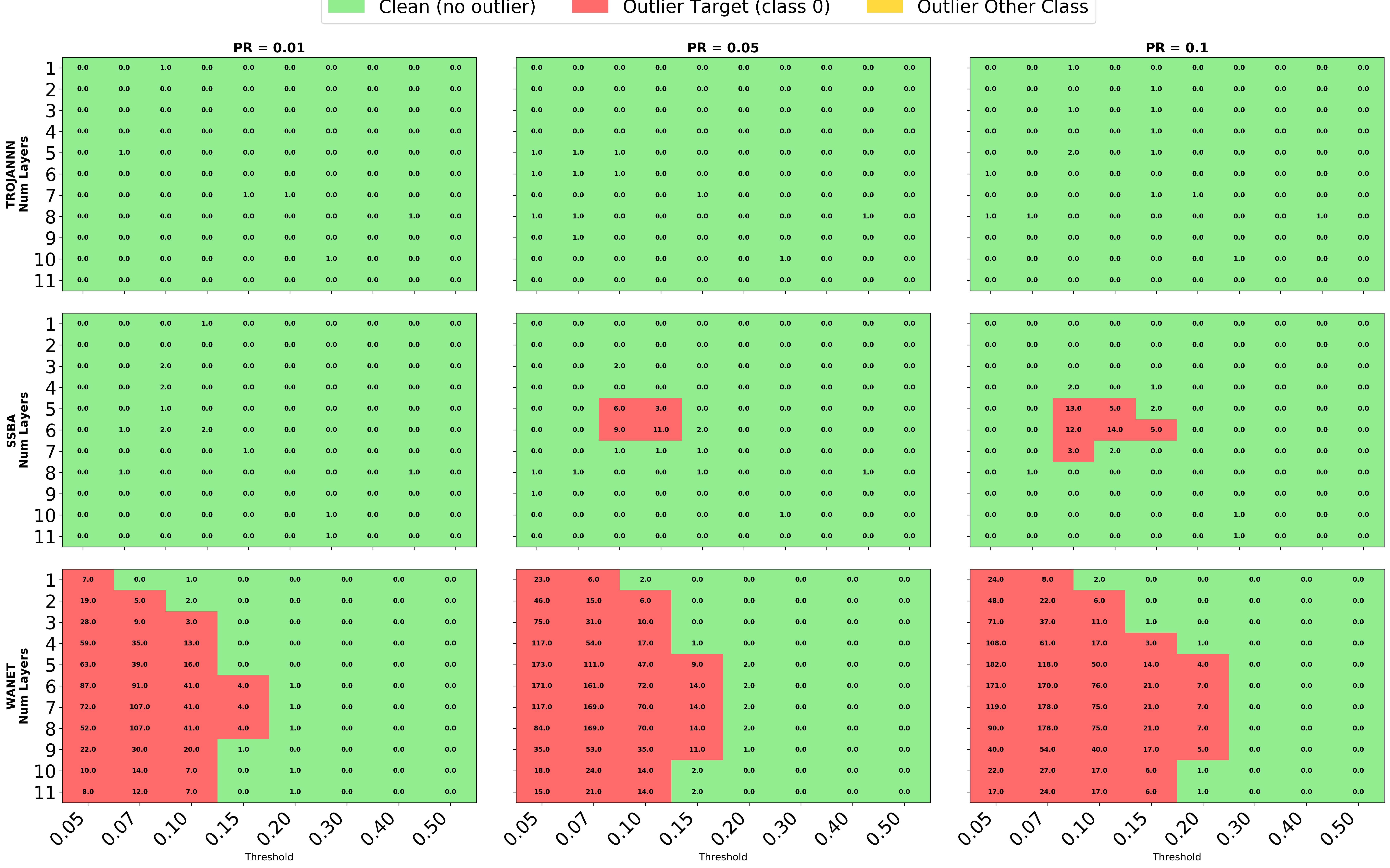}
    \caption{$Z$-scores for Tiny-ImageNet dataset across different attacks and poisoning rates.}
    \label{fig:tinydetect}
\end{figure}

In Figures~\ref{fig:cifar10detect}, Figure~\ref{fig:cifar100detect}, and Figure~\ref{fig:tinydetect}, we show threshold-layer combination results (Z-scores) for CIFAR-10, CIFAR-100 and Tiny-ImageNet datasets, respectively. The figures show results across different attacks and poisoning rates. Here we see that the detection method works for some attacks, specifically the stealthier ones, but is not successful across all attacks and all scenarios.

